\documentclass{article}

\usepackage{microtype}
\usepackage{graphicx}
\usepackage{subcaption}
\usepackage{booktabs} %

\usepackage{hyperref}

\usepackage[accepted]{include/icml/icml2026}

\usepackage{amsmath}
\usepackage{amssymb}
\usepackage{mathtools}
\usepackage{amsthm}
\usepackage{amsfonts}
\usepackage[capitalize,noabbrev]{cleveref}

\theoremstyle{plain}

\theoremstyle{definition}

\theoremstyle{remark}

\usepackage[T1]{fontenc}
\usepackage[utf8]{inputenc}
\usepackage{url}
\usepackage{nicefrac}
\usepackage{xcolor}
\usepackage{enumitem}
\usepackage{wrapfig}
\usepackage{listings}
\usepackage{framed}
\usepackage{caption}
\usepackage{array}
\usepackage{multirow}

\usepackage{tcolorbox}
\tcbuselibrary{skins}
\usepackage{afterpage}
\definecolor{linen}{RGB}{250,240,230}
\usepackage{fontawesome5}

\definecolor{movedcolor}{RGB}{0,0,0}      %
\definecolor{addedcolor}{RGB}{0,0,0}     %
\newcommand{\moved}[1]{{\color{movedcolor}#1}}
\newcommand{\added}[1]{{\color{addedcolor}#1}}

\definecolor{mydarkblue}{rgb}{0,0.08,0.45}

\hypersetup{ %
    pdftitle={Embedding Trust: Semantic Isotropy Predicts Nonfactuality in Long-Form Text Generation},
    pdfauthor={Dhrupad Bhardwaj, Julia Kempe, Tim G. J. Rudner},
    pdfsubject={},
    pdfkeywords={},
    pdfborder=0 0 0,
    pdfpagemode=UseNone,
    colorlinks=true,
    linkcolor=mydarkblue,
    citecolor=mydarkblue,
    filecolor=mydarkblue,
    urlcolor=mydarkblue,
    pdfview=FitH
}

\crefname{appsec}{appendix}{appendices}
\Crefname{appsec}{Appendix}{Appendices}

\usepackage[labelfont=bf,font=footnotesize,width=.9\textwidth]{caption}
\usepackage[format=hang]{subcaption}

\icmltitlerunning{Embedding Trust: Semantic Isotropy Predicts Nonfactuality}

\begin{document}

\twocolumn[
\icmltitle{Embedding Trust:\\Semantic Isotropy Predicts Nonfactuality in Long-Form Text Generation}

\icmlsetsymbol{equal}{*}

\begin{icmlauthorlist}
\icmlauthor{Dhrupad Bhardwaj}{nyu}
\qquad
~~~~\icmlauthor{Julia Kempe}{equal,nyu}
\qquad
\icmlauthor{Tim G. J. Rudner}{equal,toronto,vijil}~~
\end{icmlauthorlist}

\icmlaffiliation{nyu}{New York University}
\icmlaffiliation{toronto}{University of Toronto}
\icmlaffiliation{vijil}{Vijil}

\icmlcorrespondingauthor{Dhrupad Bhardwaj}{db4045@nyu.edu}
\icmlcorrespondingauthor{Tim G. J. Rudner}{tim.rudner@utoronto.ca}

\vspace*{12pt}
{\centering
  \href{https://timrudner.github.io/semantic-isotropy/}{\faIcon{globe}\, Website}
  ~~~~~
  \href{https://github.com/dhrupadb/semantic_isotropy}{\faIcon{github}\, Code}~~~~~~
\par}

\icmlkeywords{Large Language Models, Uncertainty Quantification, Factuality, Text Embeddings, Hallucination Detection}

\vskip 0.3in]

\printAffiliationsAndNotice{\icmlEqualContribution}

\begin{abstract}
To deploy large language models (LLMs) in high-stakes application domains that require substantively accurate responses to open-ended prompts, we need reliable, computationally inexpensive methods that assess the trustworthiness of long-form responses generated by LLMs. However, existing approaches often rely on claim-by-claim fact-checking, which is computationally expensive and brittle in long-form responses to open-ended prompts. In this work, we introduce semantic isotropy---the degree of uniformity across normalized text embeddings on the unit sphere---and use it to assess the trustworthiness of long-form responses generated by LLMs. To do so, we generate several long-form responses, embed them, and estimate the level of semantic isotropy of these responses as the angular dispersion of the embeddings on the unit sphere. We find that higher semantic isotropy---that is, greater embedding dispersion---reliably signals lower factual consistency across samples. Our approach requires no labeled data, no fine-tuning, and no hyperparameter selection, and can be used with open- or closed-weight embedding models. Across multiple domains, our method consistently outperforms existing aggregate trust signals in predicting nonfactuality using only a handful of samples\added{, offering a practical, low-cost first-pass signal that complements claim-level verification in real-world LLM workflows.}
\end{abstract}

\section{Introduction}
\label{introduction}

Large language models (LLMs) increasingly serve as front-line knowledge workers \citep{mckinsey2025Superagency}.
Yet the application of LLMs to high-stakes settings that require long-form responses to open-ended prompts is hamstrung by the fact that reliably ascertaining the trustworthiness of long-form text generated by LLMs remains challenging.

Without scalable, computationally inexpensive methods that are able to reliably indicate the level of trustworthiness of long-form responses generated by a language model, deployment to high-stakes application domains that require substantively accurate responses to open-ended prompts will remain risky.
Manual approaches are often infeasible at scale, and the standard stop-gaps---prompt engineering, system messages, and ensemble voting---remain brittle and expensive \citep{shorinwa2024survey}.
What is needed is a lightweight, model-agnostic, and data-agnostic method that is able to flag potentially untrustworthy text generations without needing to resort to slow, claim-by-claim fact-checking.

Prior work has attempted to gauge factuality in long-form natural language generation (NLG) by aligning atomic claims with a knowledge graph or by training auxiliary classifiers on annotated corpora \citep{wang2023surveyfactualitylargelanguage}.
However, these approaches fall short in three important ways.
First, they require structured references or costly ground-truth labels that may not exist in certain domains.
Second, they struggle with open-ended, multi-sentence answers where relevant facts are implicit rather than explicit \citep{kle}.
Third, their computational footprint grows rapidly with the number of sentences or claims, making them ill-suited to real-time applications \citep{liu2024uncertaintyestimationquantificationllms, Farquhar2024, zhang2024luq, manakul2023selfcheckgptzeroresourceblackboxhallucination}.

\begin{figure*}[t!]
  \centering
  \includegraphics[width=0.97\textwidth, trim={25, 5, 25, 10}, clip]
  {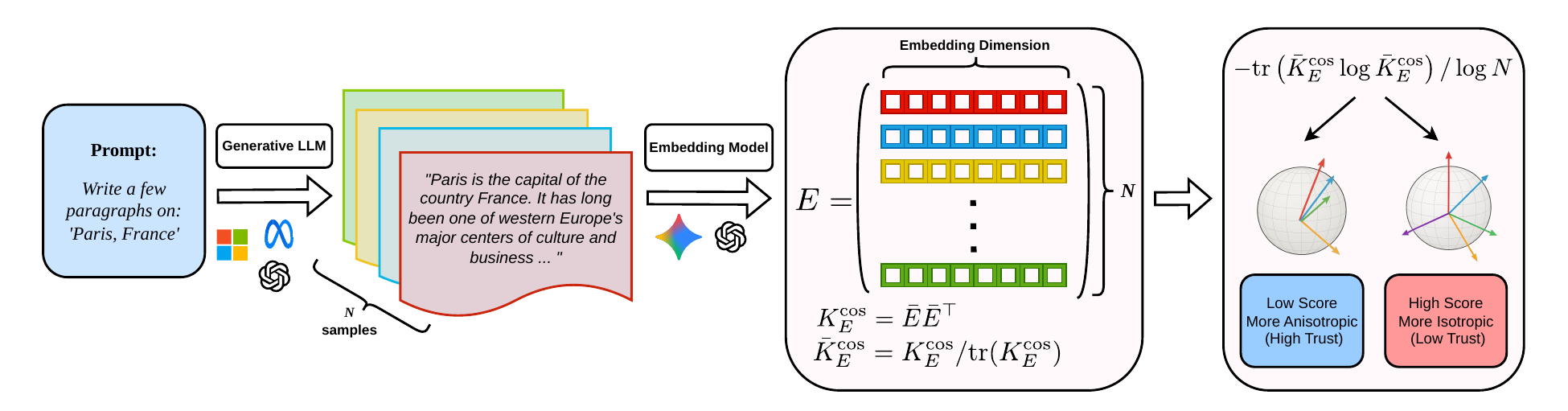}
    \caption{\textbf{Diagrammatic Overview of the {\em Semantic Isotropy} Scoring Pipeline}. This illustration presents the experimental process used to derive a {\it semantic isotropy} score.
  For a given input, we sample $N$ i.i.d. responses using a {\bf Generative LLM}.
  Each response is fed through an {\bf Embedding Model} which produces a vector representation in $R^{D}$.
  The matrix of $N\times D$ response embedding vectors is denoted as $E$.
  $E$ is transformed into a similarity matrix $K_E^{\textrm{cos}}$ using the cosine kernel, which is used to compute the isotropy score as $ \mathcal I (K_E^{\cos}) = \textrm{vNE}(K_E^{\cos})/\log N$.}
\end{figure*}

We address these limitations by introducing {\it semantic isotropy}---the degree of uniformity across normalized text embeddings on the unit sphere---and use it as a proxy for nonfactuality in long-form text generation.
Intuitively, if a prompt admits a single, factually grounded explanation, independently sampled responses from a model should cluster tightly in embedding space \citep{qiu2024semanticdensityuncertaintyquantification, wang2024subjectiveuncertaintyquantificationcalibration}.
Conversely, when an LLM hallucinates, subtle changes in semantics and content pull the embeddings apart, inflating angular dispersion.

To measure the semantic isotropy of a set of embeddings---and gauge the trustworthiness of the corresponding long-form text generations---we compute a {\it semantic isotropy score} according to the following simple and computationally inexpensive steps:
We first generate a handful of responses, we then embed them with any off-the-shelf text encoder, and finally, we compute a semantic isotropy score by estimating the von Neumann entropy of the cosine kernel under the set of embeddings.
No labels, fine-tuning, or hyperparameter search are required.

To enable broad evaluation of semantic isotropy as a proxy for nonfactuality---particularly across varying text lengths and at scale---we develop {\it Segment-Score}, a new factuality scoring method.
{\it Segment-Score} is designed to address key limitations of existing approaches \citep{min2023factscorefinegrainedatomicevaluation}:
It is more efficient in terms of token usage, scales effectively to longer responses, and offers clearer, more consistent criteria for labeling statements as true or false.
Using {\em Segment-Score}, we assess the reliability of {\em semantic isotropy scoring} in predicting nonfactuality in long-form text generation.
Computing semantic isotropy scores is model-agnostic and can be done with off-the-shelf embedding models, making this approach inexpensive and easy to use in practice.

In our empirical evaluation, we find that {\em semantic isotropy scoring} achieves state-of-the-art performance across relevant benchmarks.
It does so robustly across a wide range of models, response lengths, and evaluation settings, demonstrating its effectiveness and generalizability as a lightweight and scalable proxy for trustworthiness.

To summarize, our key contributions are:\vspace*{-3pt}
\begin{enumerate}[topsep=0pt, align=left, leftmargin=15pt, labelindent=1pt,
listparindent=\parindent, labelwidth=0pt, itemindent=!, itemsep=3pt, parsep=0pt]
    \item
    We introduce {\it semantic isotropy} and describe a simple  and computationally inexpensive method for {\it semantic isotropy scoring} as a means to assess nonfactuality in long-form natural language generation.
    \item
    We develop the {\it Segment-Score} protocol to generate and score datasets of long-form LLM generations for open-ended prompts and create a dataset of $1,182$ unique entities along with $\approx\hspace*{-3pt}65,450$ scored responses by three different models, each containing between 25 and 60 distinct claims.
    \item
    We demonstrate the efficacy of using \textit{semantic isotropy scores} as a proxy for factual inconsistency in long-form natural language generation across a range of generative models, embeddings, and relevant experimental settings.
\end{enumerate}

\vspace*{3pt}
\begin{tcolorbox}[
  enhanced,
  colback=blue!4,
  colframe=blue!55!black,
  leftrule=3pt,
  toprule=0.4pt,
  bottomrule=0.4pt,
  rightrule=0.4pt,
  arc=2pt,
  boxsep=1.5pt,
  left=8pt, right=8pt, top=4pt, bottom=4pt,
  width=\linewidth,
  before skip=2pt,
  after skip=5pt,
]
\centering
{Code to reproduce our results is available at:}
\hspace*{-4pt}\href{https://github.com/dhrupadb/semantic_isotropy}{\small \texttt{github.com/dhrupadb/semantic\_isotropy}}
\end{tcolorbox}

\section{Related Work}
\label{related_work}

\subsection{Uncertainty Quantification in Language Models}

Early work on uncertainty quantification in LLMs largely involved analyzing token-level logits and derived distributions. \citet{openai2023gpt4} demonstrated the differences in token-level calibration dynamics of pre-trained and RLHF post-trained LLMs with \citet{kirk2024} inferring that the latter process led to model miscalibration.

On the other hand, Monte Carlo sampling-based approaches \citep{zhang2020modernmontecarlomethods, malinin2021uncertaintyestimationautoregressivestructured, lin2024generatingconfidenceuncertaintyquantification, lakshmiDeepEns} have shown promise in estimating uncertainty in LLMs.
However, a key challenge highlighted by these approaches is the need to standardize LLM responses to effectively compare them.
In multiple-choice style discrete finite class output settings, studies have used a combination of Natural Language Inference (NLI) \citep{MNLI}, oracle LLMs \citep{band2024}, and embedding models \citep{qiu2024semanticdensityuncertaintyquantification, grewal2025improving} to standardize model responses for precise class attribution.
However, in NLI methods specifically, models have historically had limited context windows and several of the comparison algorithms grow quadratically in the number of inputs \citep{Farquhar2024, zhang2024luq, manakul2023selfcheckgptzeroresourceblackboxhallucination, wang2024clueconceptleveluncertaintyestimation,chen2025uncertaintyquantificationlargelanguage}. As a result, while the techniques used are in principle capable of addressing the semantic variations introduced by short-form responses, several limitations remain that hinder widespread adoption in practical settings.

A direction that addresses some of the pain points of NLI methods is the introspection of the model's internal state to estimate uncertainty. \citet{kadavath} use the activations in the model's last hidden layer to train a linear classifier probe that indicates whether the model knows the answer to the prompt. \citet{entropyprobe} adopt a similar approach by developing a cheap predictive model on top of the LLM's internal state that is capable of estimating semantic entropy \citep{Farquhar2024} without repeated sampling; \citet{kle} extend this approach to analyze a graph representation and its Laplacian. \citet{ji2024llminternalstatesreveal} use the state information to specifically predict the likelihood of hallucination while \citet{ao2024csscontrastivesemanticsimilarity} use latent representations in a multi-modal context to develop an  uncertainty quantification method for both text and images simultaneously. While the success of these approaches justifies the value of leveraging latent representations for uncertainty quantification, these existing methods generally do not scale well with increases in response size and model complexity.

\begin{algorithm*}[t!]
  \caption{\textsc{Segment-Score}($t$)}
  \label{alg:segment_verify}
  \begin{algorithmic}[1]
    \REQUIRE Topic $t$, LLM response $Y(t)$, reference document $\mathcal{D}(t)$, oracle LLM $\mathcal{O}$
    \STATE $S(t) \gets \textsf{Segmenter}_{\mathcal{O}}\bigl(Y(t)\bigr) = \langle s_1, s_2, \dots, s_m\rangle$, \quad where $m \gets |S(t)|$ \COMMENT{Partition $Y(t)$ into $m$ atomic segments}
    \STATE Initialize $\mathbf{v} \gets []$
    \FOR{$i = 1$ \textbf{to} $m$}
      \STATE $C_i \gets \langle s_1, s_2, \dots, s_{i-1}\rangle$
      \COMMENT{Context before segment $s_i$}
      \STATE $v_i \gets \textsf{Score}_{\mathcal{O}}\bigl(C_i, s_i, \mathcal{D}(t)\bigr)$
      \COMMENT{Verify $s_i$ against $\mathcal{D}(t)$ in context}
      \STATE Append $v_i$ to $\mathbf{v}$
    \ENDFOR
    \STATE $\phi \gets \tfrac{1}{m} \sum_{i=1}^m \mathbf{1}(v_i = \texttt{True})$
    \COMMENT{Fraction of segments verified as True}
    \STATE \textbf{return} $\phi$
  \end{algorithmic}
\end{algorithm*}
\vspace{-8pt} 

Recently, several studies have made meaningful advances in addressing uncertainty quantification, specifically for longer-format settings. Long Uncertainty Quantification (LUQ) \citep{zhang2024luq} and INSIDE (EigenScore) \citep{chen2024inside} are most relevant to our method and serve as key benchmarks in our analysis.
While LUQ uses NLI methods to calculate a similarity matrix between samples, the LUQ-Atomic variant that we benchmark against averages normalized entailment at a sentence / statement level, providing a much more fine-grained and robust measure of similarity.
EigenScore also samples a batch of responses and computes the eigen-distribution over the batch of latent representations in the generating model. \added{Kernel Language Entropy \citep[KLE;][]{kle} likewise uses von Neumann entropy of a kernel matrix, but its kernel is constructed from pairwise NLI entailment between sampled responses; in contrast, our cosine kernel operates directly on text embeddings, which removes the NLI bottleneck and is trivially vectorized at scale.}

\vspace*{-2pt}
\subsection{Factuality and Long-Form NLG}
\vspace*{-2pt}

While several NLP datasets are designed for multiple choice-style tasks \citep{joshi-etal-2017-triviaqa, rajpurkar2016squad100000questionsmachine, lin2022truthfulqameasuringmodelsmimic, reddy2019coqaconversationalquestionanswering}, the options for open-ended long-form responses are much more limited.
A key challenge is defining a suitable metric that is a meaningful measure of quality but also tractable to compute. Factuality, or the degree to which a response is supported by a ground truth, has been a go-to metric for several studies \citep{10.1162/tacl_a_00454}. \citet{min2023factscorefinegrainedatomicevaluation} developed FactScore, a popular scoring algorithm for generating factuality scores using a ground truth text. FactScore operates by decomposing a text into a set of atomic facts by using an oracle model (LLM) and scoring each of these facts with the help of the ground truth reference document. This method has been applied to create datasets such as FactScore-Bio \citep{zhang2024luq}, which operates on biographies taken from Wikipedia.

FactScore has given rise to several derivative methods such as LongFact \& SAFE \citep{wei2024longformfactualitylargelanguage} (which uses dynamic web searches to find supporting references), FELM \citep{chen2023felm} (applied to Math and Reasoning problems), Multi-FAct \citep{shafayat2024multifactassessingfactualitymultilingual} (multi-lingual setting), FactAlign \citep{huang2024factalignlongformfactualityalignment} and LoGU \citep{yang2024logulongformgenerationuncertainty} (post-training / RLHF to incorporate uncertainty phrasing into model generations), and VeriScore \citep{song2024veriscoreevaluatingfactualityverifiable} (better precision by scoring verifiable facts only) to name a few. But most of these methods share key limitations of the FactScore framework, scaling poorly with response length and requiring several LLM calls per sample. We address some of these drawbacks through our method, \textit{Segment-Score} (see Algorithm \ref{alg:segment_verify}), which utilizes capabilities of more modern LLMs such as longer context windows and the ability to respond with structured outputs.

\vspace*{-5pt}
\section{\mbox{Semantic\hspace*{1.7pt}Isotropy\hspace*{1.7pt}in\hspace*{1.7pt}Long-Form\hspace*{1.7pt}Generation}}
\label{method}
\vspace*{-3pt}

In this section, we introduce {\it semantic isotropy}, a measure of semantic variation in long-form text generations, and use it to motivate  {\it semantic isotropy scoring}, a method that expresses the level of semantic isotropy in long-form responses and can serve as a proxy for nonfactuality.

\vspace*{-1pt}
\subsection{Semantic Isotropy}
\vspace*{-1pt}

We start with the conventional definition of isotropy.
Intuitively, isotropy describes a state where a set of vectors or a distribution has no preferred direction---its properties are the same in all orientations.
More formally, for a set of unit vectors, we can express this property as follows:\vspace*{2pt}
\begin{tcolorbox}[
  enhanced,
  colback=blue!4,
  colframe=blue!55!black,
  leftrule=3pt,
  toprule=0.4pt,
  bottomrule=0.4pt,
  rightrule=0.4pt,
  arc=2pt,
  boxsep=1.5pt,
  left=8pt, right=8pt, top=4pt, bottom=4pt,
  width=\linewidth,
  before skip=2pt,
  after skip=5pt,
]
{\bf Definition (Isotropy).}
{\it
    The column vectors ${x_{1}, ..., x_{N} \in \mathbb{R}^D}$ with \mbox{$\left\|x_{i}\right\|_2=1$} for all \mbox{$i \in \{ 1, ..., N \}$} are said to be isotropic if for the matrix
    ${X = [x_{1}, ..., x_{N}]}$:  ${X^{\top} X}=I_N$.
}
\end{tcolorbox}
\vspace*{1pt}
This condition means that, on average, the transformation represented by $X$ preserves directions and spreads uniformly across all directions in $\mathbb{R}^D$. When $X^{\top}X = I_N$, the linear transformation $x \rightarrow X^{\top}x$, for $x \in \mathbb{R}^D$, acts as an isometry on $\mathbb{R}^N$ embedded in $\mathbb{R}^D$ (with $N \ll D$ in our case). Geometrically, if the unit vectors $x_{1}, ..., x_{N}$ are isotropic, this implies that the unit vectors are uniformly dispersed across the unit sphere, giving us an intuitively meaningful measure of variation within $x_{1}, ..., x_{N}$.

We will now build on this intuition to define {\it semantic isotropy}.
Let $\mathcal{R} = \{ r_{1}, ..., r_{N} \} \in \mathcal{T}^{N} $ be a set of $N$ long-form NLG responses given a prompt $p$ and let $\Gamma_{\theta}(\cdot) : \mathcal{T} \to \mathbb{R}^D$ be an embedding model parameterized by neural network parameters $\theta$.
We define $E=\{e_1, ... ,e_N\}$ to be the collection of embedding vectors, where $e_{i} = \Gamma_{\theta}(r_{i}) \in \mathbb{R}^{D}$ is the embedding vector of response $r_{i}$,  $\bar{e}_{i} \in \mathbb{R}^{D}$ is the corresponding normalized embedding vector, $\bar{e}_{i} = e_{i}/ ||e_{i}|| $,
and $\bar{E}$ is the matrix of normalized embedding vectors, $\bar{E} = [\bar{e}_{1}, ..., \bar{e}_{N}]^\top \in \mathbb{R}^{N \times D}$.
The cosine kernel function for two embedding vectors, $e_{i}, e_{j} \in \mathbb{R}^{D}$, is then defined as
\begin{align}
     k^{\textrm{cos}}(e_{i}, e_{j}) = [ \bar{E} \bar{E}^{\top} ]_{i,j} = \bar{e}_{i}^\top \bar{e}_{j} .
\end{align}
We denote the cosine kernel matrix for $E$ by $K_E^{\textrm{cos}} \in \mathbb{R}^{N \times N}$.

With these preliminaries in place, we are now ready to formally define {\it semantic isotropy} to represent uniform dispersion of the normalized embedding vectors:\vspace*{1pt}
\vspace*{2pt}
\begin{tcolorbox}[
  enhanced,
  colback=blue!4,
  colframe=blue!55!black,
  leftrule=3pt,
  toprule=0.4pt,
  bottomrule=0.4pt,
  rightrule=0.4pt,
  arc=2pt,
  boxsep=1.5pt,
  left=8pt, right=8pt, top=4pt, bottom=4pt,
  width=\linewidth,
  before skip=2pt,
  after skip=5pt,
]
{\bf Definition (Semantic Isotropy).}
{\it
    A set $\smash{\mathcal{R} = \{ r_{1}, ..., r_{N} \}}$  of long-form responses with corresponding embeddings $\smash{E=\{e_{1}, ..., e_{N}\}}$ is semantically isotropic if $K_E^{\textrm{\em cos}}=I_N$.
}
\end{tcolorbox}
\vspace*{1pt}

\subsection{Measuring Semantic Isotropy in Long-Form Language Generation}

As defined, isotropy represents a strict condition that is either met or not met.
However, to use semantic isotropy in practice as a means of ascertaining nonfactuality in long-form language generation, we need to gauge the {\it level} of isotropy of a given cosine kernel.
We want to formalize the intuition that factuality corresponds to a high level of semantic alignment within a set of long-form generations (highly anisotropic), whereas uncertainty and nonfactuality correspond to a high level of dispersion (very isotropic).

\begin{figure*}[t!]
  \centering
  \includegraphics[width=0.95\textwidth,trim={0 30 0 29},clip]{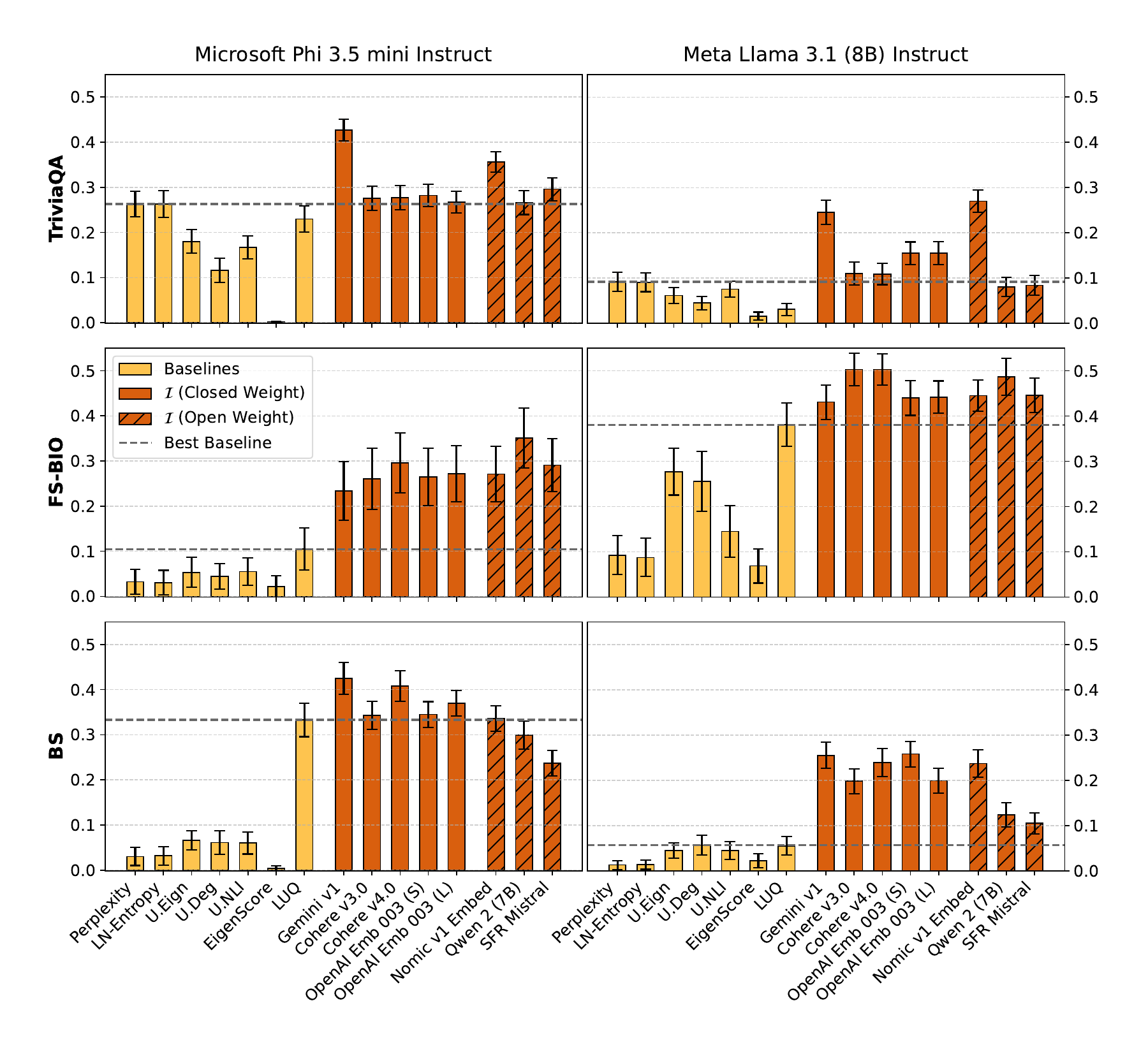}
  \caption{\textbf{Main Experimental Results}. Bar charts comparing the performance (as measured by the $R^2$ of a linear model of Factuality $\sim$ Semantic Isotropy); implemented using various embedding models and benchmark uncertainty metrics on three datasets: TriviaQA Entities \textbf{[TriviaQA]} (Top Row), FactScore-Biographies \textbf{[FS-BIO]} (Middle Row) and CMU Book Summaries \textbf{[BS]} (Bottom Row); scored using the \textit{Segment-Score} \textbf{(SS)} algorithm. Responses are $\approx $500 words. 1,500 bootstrapped samples are used to generate 1-SD error bars. Semantic isotropy $\mathcal{I}$ (ours) outperforms all other baselines for all embedding models.}
\label{fig:main_results}
\end{figure*}

\begin{figure*}[t!]
  \centering
  \includegraphics[width=0.95\textwidth,trim={0 25 0 15},clip]{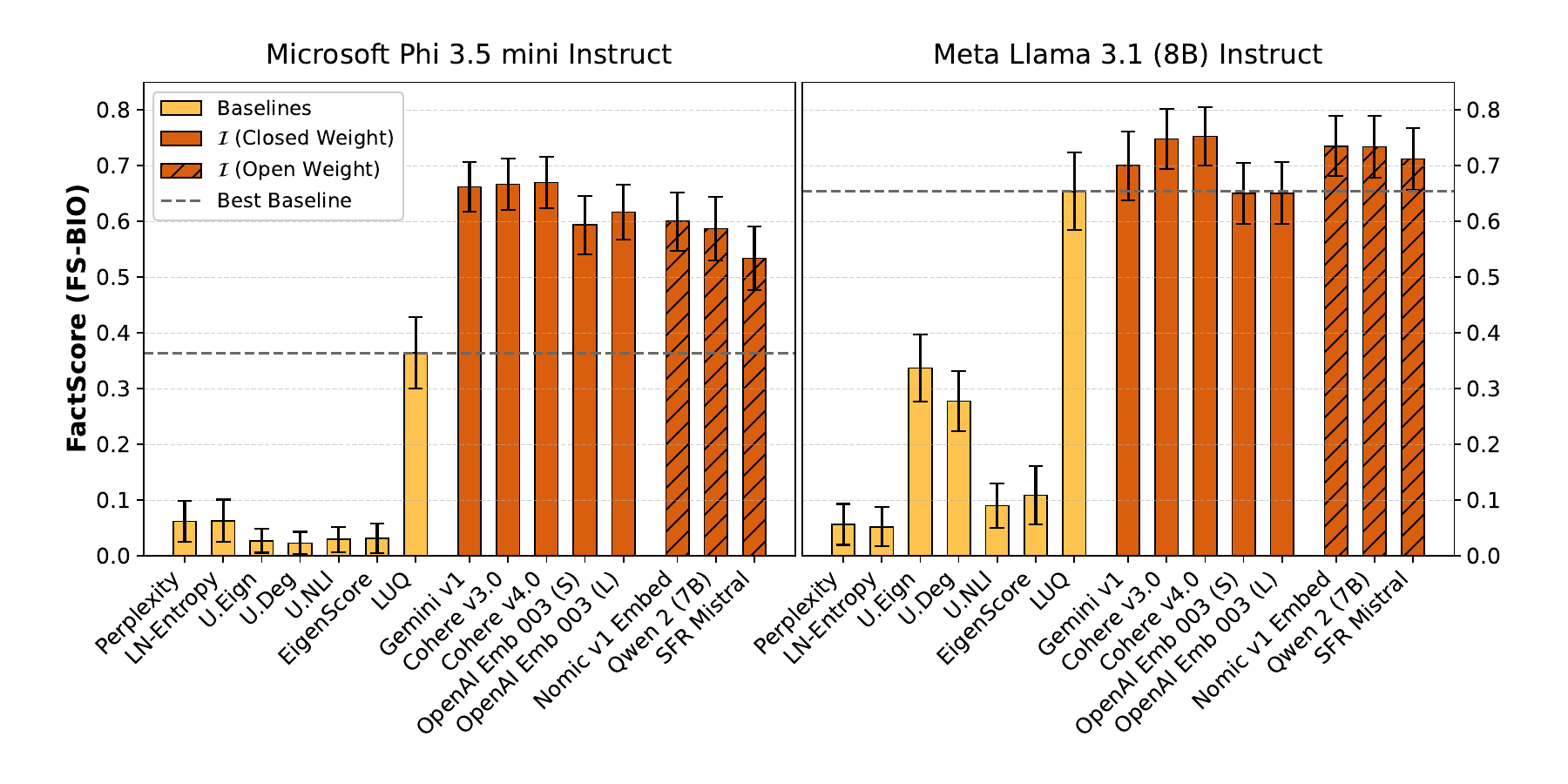}
    \caption{\textbf{FactScore Experimental Results}. Bar charts comparing the performance (as measured by the $R^2$ of a linear model of factuality with semantic isotropy as the explanatory variable); implemented using various embedding models and benchmark uncertainty metrics on the \textbf{FS-BIO} dataset using the \textit{FactScore} \textbf{(FS)} scoring algorithm.
    Left hand side: Benchmark UQ metrics. Right Hand Side: semantic isotropy  (\textbf{$\mathcal{I}$}) implemented using various embedding models.
    Same experimental setting as in \Cref{fig:main_results}. We observe that our method's performance is robust to the scoring scheme.}
    \label{fig:factscore_fsbio_results}
\end{figure*}

To estimate the level of isotropy of a set of long-form text generations, we use the von Neumann entropy (vNE) \citep{math_quantum_foundations} of the cosine kernel $\smash{K_E^{\cos}}$, normalized to have $\textrm{trace}=1$, which allows us to view the eigenvalues as defining a discrete probability distribution.
Let $\smash{\bar K_E^{\cos} =K_E^{\cos}/\textrm{trace}(K_E^{\cos})}$ and define
\begin{align}
    \textrm{vNE}(K_E^{\cos})
    =
    -\textrm{trace}({\bar K_E^{\cos}} \log \bar K_E^{\cos}) . \label{eqn:vn_matrix}
\end{align}
To be precise, we view the above definition in the eigenspace of the positive semidefinite matrix $\smash{K_E^{\cos}}$, that is, $\smash{\textrm{vNE}(K_E^{\cos}) = - \sum_{i}^N\lambda_{i} \log \lambda_{i}}$, where $\lambda_1 ,..., \lambda_N$ are the eigenvalues of $\smash{\bar K_E^{\cos}}$.
Note that $\textrm{vNE}(K_E^{\cos}$) attains its maximum value of $\log N$ for semantically (perfectly) isotropic responses (i.e., isotropic normalized embedding vectors) and is minimized at $0$ when all embedding vectors are aligned (parallel).
It thus measures the degree of dispersion of $\smash{\{\bar{e}_{i}\}_{i=1}^{N}}$ on the unit sphere and serves as a measure of the level of isotropy within the set of text embeddings.

We are now ready to define the {\em semantic isotropy score} $\mathcal I(\cdot)$ that will serve as our proxy for factuality.
Consider $N$ sampled long-form text generations in response to a prompt and define\vspace*{-3pt}
\begin{align}
    \mathcal I (K_E^{\cos}) = \textrm{vNE}(K_E^{\cos})/\log N,
\end{align}%
with $0 \leq \mathcal I (K_E^{\cos}) \leq 1$.
Under this definition, (i) a lower degree of semantic isotropy (a semantic isotropy score closer to $0$) is associated with stronger alignment/lower dispersion and consequently higher certainty and factuality, whereas (ii) a higher degree of semantic isotropy (a semantic isotropy score closer to $1$) is associated with weaker alignment/higher dispersion and consequently lower certainty and factuality.
In a nutshell: {\it The lower} the semantic isotropy score, {\it the more trustworthy} the generating model's long-form responses to the prompt are.

\section{Empirical Evaluation}
\label{results}

In this section, we conduct an empirical evaluation to assess whether semantic isotropy scores can serve as an effective proxy for nonfactuality in long-form natural language generation with LLMs.
First, we will describe the creation of a ground truth evaluation dataset using {\em Segment-Score} and present our evaluation metrics and benchmarks.
We will then present our main findings about the efficacy of semantic isotropy scoring as a means to predict nonfactuality in long-form text generation.

\subsection{Creation of a Dataset of Long-Form Responses to Open-Ended Prompts}
\label{sec:dataset_generation}

To construct a dataset of long-form responses to open-ended prompts, we must select a corpus of entities such that each is associated with an underlying ground truth reference document.
These entities are used to create a set of open-ended prompts that can be fed into an LLM to generate long-form responses.
We use three sources---\textit{FactScore-Bio} [\textbf{FS-BIO}] (182 unique entities) \citep{min2023factscorefinegrainedatomicevaluation, zhang2024luq},  \textbf{TriviaQA} \citep{joshi-etal-2017-triviaqa} (1,000 unique entities) and CMU Book Summaries [\textbf{BS}] (509 unique entities) \citep{bamman2013newalignmentmethodsdiscriminative}.
All three datasets use Wikipedia as their ground truth to source reference documents.
For \textbf{TriviaQA}, we exclude all entities that correspond to days, dates or numerical values and only select those that match the title of the underlying Wikipedia page.
Of the 5,245 qualifying TriviaQA entities across the training and validation set, we randomly select a subset of 1,000 entities to create open-ended prompts for long-form text generation. For \textit{Book Summaries} (\textbf{BS}), we constrain the dataset to entities where the underlying reference text is at least 5,000 characters and the publish date is known, and is on or after Jan 1, 2000. Of a total of 16,559 entries, we select 509 entities that meet this criterion.

To generate long-form responses, we use Meta Llama 3.1 8B Instruct \citep{grattafiori2024llama3herdmodels}, Microsoft Phi 3.5 Mini Instruct \citep{abdin2024phi3technicalreporthighly}, and OpenAI GPT 4.1 Mini \citep{gpt41} to generate responses for \textbf{TriviaQA} and \textbf{FS-BIO}. For \textbf{BS}, we only generate responses using Llama 3.1 and Phi 3.5 Mini.
For \textbf{TriviaQA} we sample up to $k=20$ responses while for \textbf{FS-BIO} and \textbf{BS} we sample $k=10$ responses, targeting approximately $500$ words for each response.
For \textbf{TriviaQA} using Llama 3.1 8B Instruct, we also generate a longer dataset of approximately $1,000$ words and derived datasets of intermediate word counts of approximately $125$, $250$, $375$, $500$, and $750$ words by truncating the $1,000$ word variant to the nearest sentence.
Model inference is run with temperature $\tau = 0.7$ and FP16 quantization using the vLLM framework \citep{kwon2023efficientmemorymanagementlarge}.
Finally, we compute factuality scores for the datasets using the \textit{Segment-Score} \textbf{(SS)} procedure (\Cref{alg:segment_verify}).
Implementation details and prompts are presented in \Cref{sec:segscore_method}. The dataset and Segment-Score algorithm are released to allow future benchmarking by the research community.

We also generate \textit{FactScore} \citep[\textbf{FS};][]{min2023factscorefinegrainedatomicevaluation} scores for the \textbf{FS-BIO} datasets from each model.
This allows us to benchmark our scoring algorithm \textbf{(SS)} against the existing state-of-the-art method, and we can show that our results hold robustly.
See \Cref{sec:ss_vs_fs} for a comparison of the scoring methods.
We use GPT 4.1 Mini \citep{gpt41} as the oracle LLM in both algorithms.

\begin{figure*}
  \centering
  \includegraphics[width=0.95\textwidth,trim={0 15 0 30},clip]{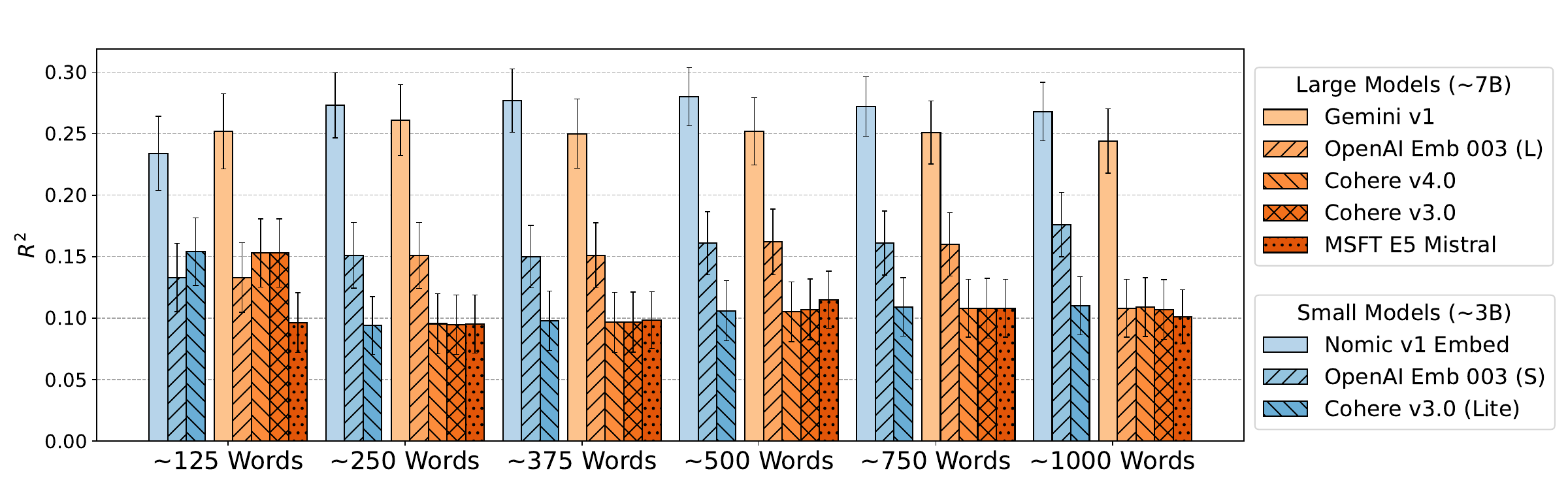}
  \caption{\textbf{Performance by Response Length}: Comparing \textit{Semantic Isotropy} in terms of explained variance ($R^2$) of Factuality \textbf{(SS)} across various response lengths, contrasted by model size: Small (up to $3B$ parameters) and Large ($\approx\hspace*{-3pt}7B$ or more parameters). 1,500 bootstrapped samples are used to generate 1-SD error bars.}
  \label{fig:response_length_analysis}
\end{figure*}

\subsection{Evaluation Metrics}
\label{sec:eval_metrics}

To assess the performance of our methods and compare them to appropriate baselines from the literature, we measure the degree of the relationship between each individual measure and the average factuality of a given topic, aggregated across all responses.
First, we selected a set of embedding models that rank highly on the MTEB Benchmark \citep{mteb} to use in computing our isotropy scores.
We measure predictiveness of these scores using $R^2$, the explained variance of a linear model with factuality scores as the dependent variable and isotropy scores as the independent variable.
To estimate error bounds, we perform bootstrap sampling and generate 1-SD error bars for each measure.
We compare \textit{Semantic Isotropy} to the following baseline measures:
\textbf{Perplexity} - Exponential of the average negative log-likelihood of the generation \citep{ren2023outofdistributiondetectionselectivegeneration}; \textbf{LN-Entropy} - Length Normalized Entropy \citep{malinin2021uncertaintyestimationautoregressivestructured}; \textbf{U.Eign} - Sum of Eigenvalues of the Graph Laplacian, \textbf{U.Deg} -  Average pairwise distance measured using an NLI model \citep{lin2024generatingconfidenceuncertaintyquantification}; \textbf{U.NLI} - NLI uncertainty derived using SelfCheckNLI \citep{manakul2023selfcheckgptzeroresourceblackboxhallucination}; \textbf{Semantic Entropy} - Entropy of the probability distribution of semantic clusters over the responses \citep{Farquhar2024}; \textbf{LUQ-Atomic} - Long Uncertainty Quantification using atomic facts via LLM based fact extraction \citep{zhang2024luq};  \textbf{EigenScore} - the log-determinant of the centered embedding matrix of the generating model's internal state activations over a set of sampled responses \cite{chen2024inside}. \added{Each baseline method is implemented using the original backbone (i.e., DeBERTa-v3 Large (MNLI) for the NLI components of LUQ, SelfCheckNLI, and semantic entropy; internal-state activations for EigenScore); full numerical results, including semantic entropy, are reported in \Cref{tab:main_results_table_ss,tab:main_results_table_fs} (\Cref{sec:appendix:full_results}).}

While relevant to our study, we do not include comparisons with \textbf{Kernel Language Entropy (KLE)} \citep{kle} and \textbf{Graph Longform Uncertainty} \citep{graph_longform_uncertainty}. We found that the KLE computation was highly numerically unstable given the degree of semantic entailment overlap among the sampled long-form responses and did not produce meaningful results. For Graph Uncertainty, the distinct claim union algorithm proved computationally difficult to implement at the response lengths considered in our study. For context, a typical set of responses had over 600 distinct claims, which implied 6,000 LLM calls to measure claim to response entailment for one topic, assuming $N=10$.

\paragraph{Sampling the appropriate embedding.}
Unlike closed-weight models that return just an embedding vector, for open-weight models we select the activations corresponding to the last token in the final hidden state, except for Nomic v1 \citep{nussbaum2025nomicembedtrainingreproducible} where we take the average of the activations over the token dimension instead.
For further details, see \Cref{sec:sensitivity_eval}.

\vspace{-5pt}
\subsection{Main Result: Factuality in Long-Form Responses to Open-Ended Prompts}
\label{sec:main_results}

We present our main empirical results in \Cref{fig:main_results} (see also \Cref{tab:main_results_table_ss} in \Cref{sec:appendix:full_results}).
We find that semantic isotropy scores are superior to existing methods in predicting nonfactuality as per the \textbf{(SS)} algorithm on the \textbf{FS-BIO}, \textbf{TriviaQA} and \textbf{BS} datasets.
Using $R^2$ (i.e., the explained variance) as our performance measure, $\mathcal{I}$ outperforms all existing metrics, in most cases by a wide margin.
We see some variance in the performance under different embedding models, and discuss this finding in more detail in \Cref{sec:sensitivity_eval}.
While certain embedding models are better suited to specific tasks, semantic isotropy scoring outperforms existing state-of-the-art methods across the board.
Notably, the Nomic v1  embedding \citep{nussbaum2025nomicembedtrainingreproducible} demonstrates exceptional performance on all datasets, occasionally surpassing larger and closed-source models such as Gemini \citep{geminiteam2024geminifamilyhighlycapable} and OpenAI Embeddings \citep{gptembed}.

\afterpage{%
\begin{figure}[t!]
        \centering
        \includegraphics[width=0.9\linewidth,trim={40 30 30 30},clip]{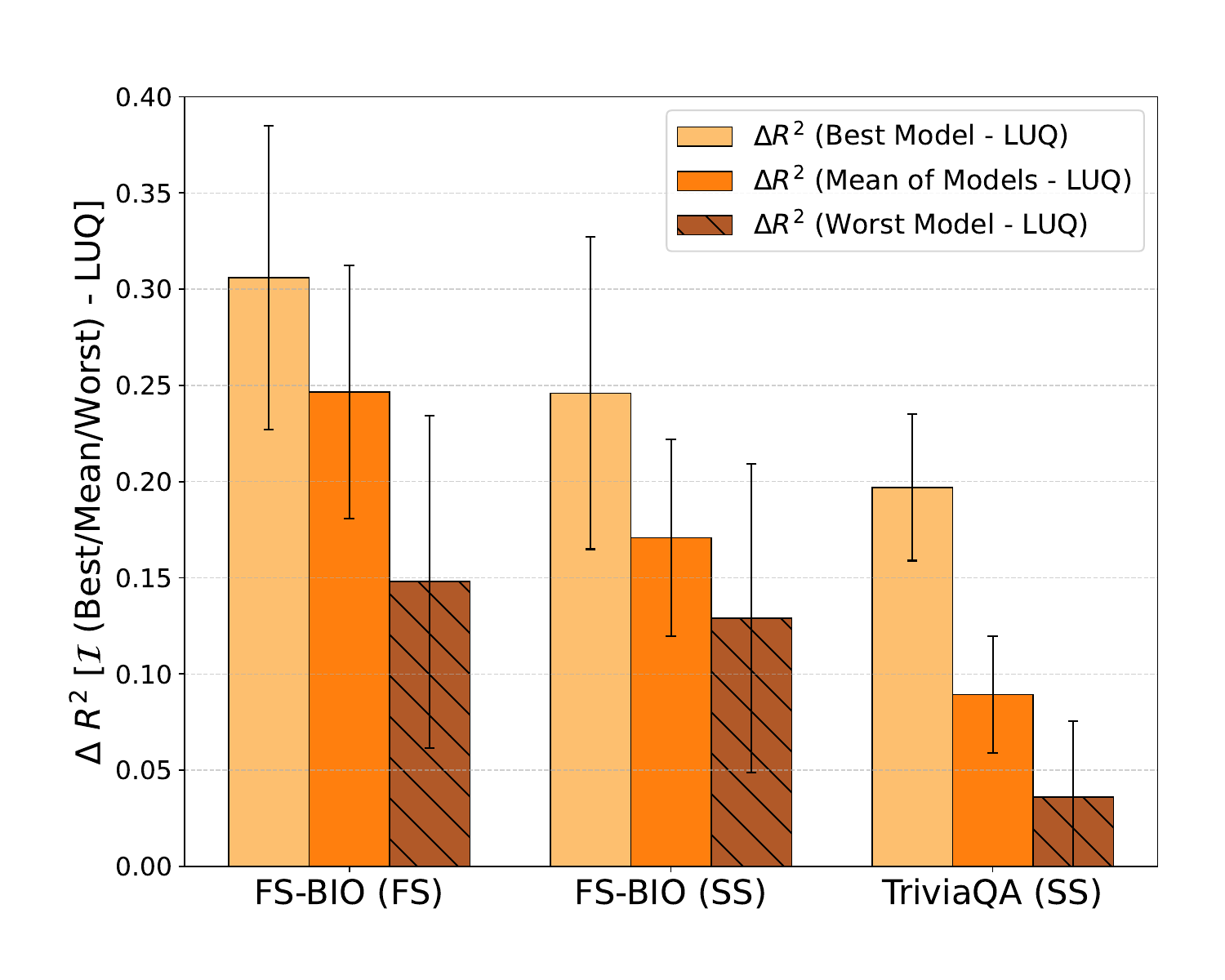}
        \captionsetup{width=.95\linewidth}
        \caption{\textbf{Model Agnostic Performance Comparison}: Performance difference between current state-of-the-art \textbf{LUQ-Atomic} \citep{zhang2024luq} and semantic isotropy based on different embedding models across each dataset using Phi 3.5 Mini Instruct as the generator model. In each case, all versions of semantic isotropy are more predictive than LUQ. 1,500 bootstrapped samples are used to generate 1-SD error bars.}
        \label{fig:luq_delta_plot}
\vspace{-8pt}
\end{figure}
}

\vspace{-5pt}
\subsection{FactScore Evaluation}
\label{sec:factscore_eval}

To ensure that our results are not merely an artifact of the \textit{Segment-Score} method, we also evaluate semantic isotropy scoring under the \textit{FactScore} (\textbf{FS}) algorithm applied to the $\textbf{FS-BIO}$ dataset (see \Cref{fig:factscore_fsbio_results}).
We find that the results mirror those in \Cref{fig:main_results}, even improving the predictive performance of semantic isotropy scoring.
This consistency across scoring schemes highlights the robustness and generalizability of semantic isotropy as a measure of nonfactuality.
A more detailed comparison of the two scoring methods can be found in \Cref{sec:ss_vs_fs}.

\begin{figure*}[t!]
  \centering
  \includegraphics[width=0.95\textwidth,trim={70 30 20 29},clip]{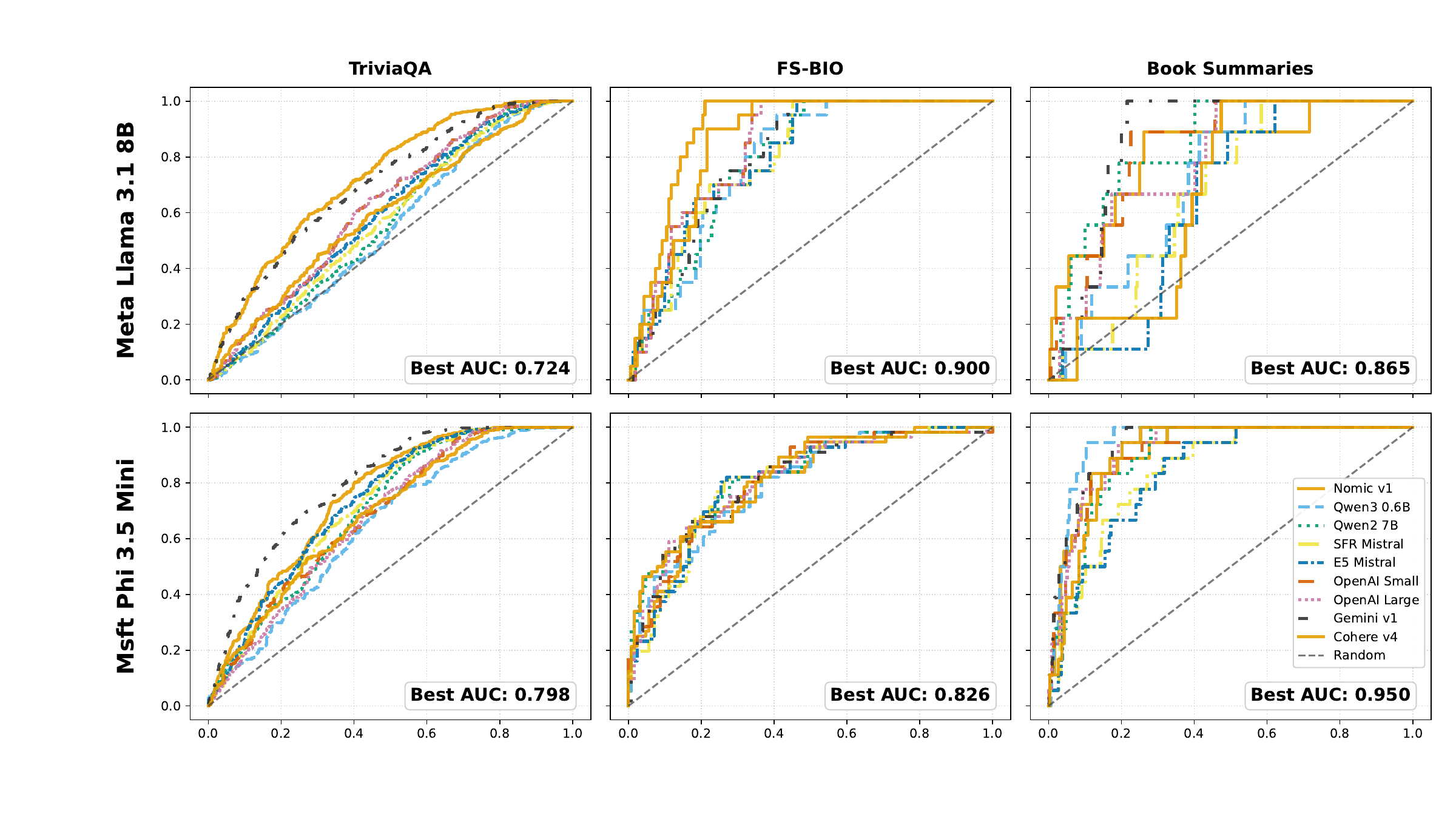}
  \vspace*{-5pt}
  \caption{\textbf{Isotropy vs.\ Factuality (ROC).} ROC curves comparing \textit{Semantic Isotropy} against $\{0,1\}$ class labels obtained by thresholding \textit{Segment-Score} at $50\%$. Curves are computed on Llama 3.1 and Phi 3.5 Mini responses across all three datasets.}
  \label{fig:auroc}
\end{figure*}

\moved{
\subsection{Discriminative Power at the Operating Point}
\label{sec:auroc}

To assess discriminative power at a concrete operating point, we discretize the factuality target: topics with average \textit{Segment-Score} factuality above $50\%$ are labeled $1$ and the remainder $0$, with $1-\mathcal{I}$ serving as the confidence score. The resulting receiver operating characteristic curves are shown in \Cref{fig:auroc}.
On the entity-grounded datasets, the score discriminates strongly, with the AUC reaching a value of approximately $0.95$ on Book Summaries (Phi 3.5 Mini) and a value of $0.90$ on FS-BIO (Llama 3.1).
High-isotropy prompts can therefore be routed to expensive claim-level verification while low-isotropy ones pass with minimal additional checking; threshold calibration is inherently deployment-specific, following \mbox{standard practice for sampling-based uncertainty scores.}
}

\moved{
\subsection{Computational Cost}
\label{sec:compute}

LUQ and Graph Uncertainty scale as $\mathcal{O}(MN^2)$, where $M$ is the average number of facts or segments per response and $N$ is the number of sampled responses, since each pair of segments requires an NLI forward pass. Semantic entropy and related methods are $\mathcal{O}(N^2)$ in entailment computations and cannot be vectorized naively. Semantic isotropy is also $\mathcal{O}(N^2)$ in the kernel construction, but the cosine kernel is fully parallelizable and the cost of the vNE computation on an $N\times N$ matrix with $N \leq 20$ is negligible.
In practice, computing isotropy scores for one batch of $N=20$ responses, each comprised of approximately $500$ words, requires $1.8 \pm 0.05$ seconds on a V100 (amortized, Nomic v1) compared to $302 \pm 48$ seconds for LUQ-Atomic using DeBERTa-v3 Large (MNLI) \citep{manakul2023selfcheckgptzeroresourceblackboxhallucination} of comparable parameter count---approximately $170\times$ faster.
}

\section{Sensitivity Studies}
\label{sec:sensitivity_eval}

\paragraph{Effect of Response Length.}
One important study we undertake is to understand the relationship between measure performance and response length.
While our main study focuses on $\approx\hspace*{-3pt}500$ word responses, we explore the relationship between semantic isotropy and factuality for shorter and longer texts as well.
We outline these results in \Cref{fig:response_length_analysis}.
While certain models perform better (Gemini and Nomic v1), performance is generally consistent across the board. OpenAI's Text Embedding 3 is a notable exception, with performance consistently increasing with response length, independent of model size.

\paragraph{Effect of Choice of Embedding Model.}
We analyze how the size and type of embedding model impact performance. As shown in \Cref{fig:luq_delta_plot}, {\em Semantic Isotropy} consistently outperforms LUQ across all three datasets/scoring methods and embedding models. Notably, even the least effective embedding model yields improvements over LUQ, while the best performs substantially better. These results highlight the robustness of {\em Semantic Isotropy}, demonstrating its effectiveness regardless of the embedding model used.
We compare small and large variants of closed source models such as those provided by OpenAI and Cohere. In both cases, the performance difference is negligible and likely attributable to sampling noise. This finding demonstrates that semantic isotropy is robust even at small model sizes. \added{Because the factual-vs.-nonfactual signal we exploit is a coarse, distribution-level effect, encoders with lower-dimensional representations \citep{kusupati2022matryoshka} retain it well, often matching larger models whose additional capacity primarily captures finer-grained stylistic detail.}

\paragraph{Effect of Choice of Isotropy Measure.}
In \Cref{tab:isotropy-measures} (\Cref{sec:appendix:full_results}) we ablate a variety of alternative measures of isotropy, among them the Frobenius norm $\|K_E^{\cos}\|$, $\log(\det(K_E^{\cos}))$ and trace$(K_E^{\cos})^{-1}$.
We find that von Neumann entropy-based isotropy scoring performs best overall, outperforming other measures on average across different embedding model choices and dataset settings.
However, most of these measures demonstrate nearly equivalent performance, further underscoring the robustness of semantic isotropy as a proxy for nonfactuality.

\begin{table}[t]
\centering
\captionsetup{width=\linewidth}
\caption{\small \textbf{Segment-Score Coherence Test}: Comparison of label agreement of \textit{Segment-Score} classifications and those obtained using a majority voting scheme. For each dataset, we randomly sample 30 entities and score their response corpus using GPT 4.1, Claude 4 Sonnet and DeepSeek, taking 5 samples each. Scores are in percentage (\%).}
\vspace{-10pt}
\label{tab:coherence}
\vspace{0.1in}
\begin{small}
\setlength{\tabcolsep}{5pt}
\begin{tabular}{llccc}
\toprule
\textbf{Dataset} & \textbf{Gen.\ Model} & \textbf{OpenAI} & \textbf{Claude} & \textbf{DeepSeek} \\
 & & \scriptsize{(GPT 4.1)} & \scriptsize{(4 Sonnet)} & \\
\midrule
\multirow{2}{*}{\textbf{BS}} & Llama 3.1 & \textbf{89.24} & 81.33 & 75.87 \\
 & Phi 3.5 Mini & \textbf{91.28} & 86.13 & 80.38 \\
\midrule
\multirow{2}{*}{\textbf{FS-BIO}} & Llama 3.1 & 83.92 & 86.13 & \textbf{86.39} \\
 & Phi 3.5 Mini & 76.98 & 86.67 & \textbf{88.30} \\
\bottomrule
\end{tabular}
\end{small}
\end{table}

\paragraph{Coherence of Segment-Score.} While \textit{Segment-Score} offers several practical benefits, it remains critical to establish that it is ultimately a sound proxy for factuality. To calibrate FactScore, \citet{min2023factscorefinegrainedatomicevaluation} used a cohort of human evaluators. Given the size and span of our datasets, we opted for an LLM-assisted workflow using several SOTA oracle models such as GPT 4.1 \citep{gpt41} or Claude 4 \citep{claude4}. For each response, we sample up to $5$ True/False labels for each segment within, given the reference document. We use the majority of these labels as our proxy for a true class label. In \Cref{tab:coherence}, we provide the matching accuracy against the one-shot labels generated by \textit{Segment-Score}. The high degree of agreement indicates that the \textit{Segment-Score} algorithm can be effectively utilized to proxy factuality. In \Cref{sec:ss_vs_fs}, we elaborate on the study relating the scores obtained by \textit{Segment-Score} and \textit{FactScore} algorithms on the \textbf{FS-BIO} dataset.

\section{Limitations}
\label{sec:limitations}

Computing semantic isotropy scores requires repeated sampling, which incurs extra inference cost relative to single-pass probe-based uncertainty estimates, but this overhead is decreasing with modern high-throughput inference engines and reasoning-oriented LLMs.
Furthermore, semantic isotropy scoring operates at the prompt level and does not localize errors to particular claims, so it is best paired with claim-level verification when fine-grained attributions are required.
\added{Semantic isotropy further relies on extrinsic divergence under uncertainty, that is, the empirical regularity that hallucinated responses tend to disagree with one another \citep{manakul2023selfcheckgptzeroresourceblackboxhallucination, ricco2025hallucinationdetectionprobabilisticframework}.
As such, the nonfactuality signal obtainable through semantic isotropy scoring weakens for generators whose outputs remain stylistically uniform even when hallucinating, as we observe with GPT 4.1 Mini on TriviaQA (\Cref{fig:openai_analysis}, with a style analysis in \Cref{fig:isotropy_kde,fig:token_count_distribution,fig:openai_comparative_analysis_summary}), and is uninformative for intrinsic hallucinations that remain self-consistent across samples.}
Across the broad range of settings considered in our empirical evaluation, however, semantic isotropy consistently identifies nonfactual long-form generations with high accuracy and provides a practical first-pass signal that can be paired with claim-level verification if needed.

\section{Conclusion}
\label{conclusion}

In this work, we introduced {\em Semantic Isotropy}, proposed a computationally inexpensive method for semantic isotropy scoring, presented {\em Segment-Score} for cheap long-form ground truth data generation, and demonstrated that semantic isotropy scores yield state-of-the-art factuality prediction in long-form natural language generation.
We found that semantic isotropy is highly predictive of nonfactuality in long-form LLM responses to open-ended prompts and that semantic-isotropy scores outperform alternative approaches---in most cases by a large margin.
We also found that, in most of the cases considered in our analysis, semantic isotropy scoring is robust to the choice of embedding model, the choice of isotropy measure, the response length, and the number of samples used.
We released code to reproduce the empirical evaluation and implement the {\em Segment-Score} method, and we will make the {\it Segment-Score}-annotated dataset publicly available to facilitate further research into assessing factuality in long-form text generation.
Our findings suggest that semantic isotropy is a simple and yet effective predictor of nonfactuality and open the door to more reliable and cost-effective evaluation of long-form LLM text generation at scale.

\clearpage

\section*{Impact Statement}

This paper presents work whose goal is to advance the field of Machine Learning by providing methods to assess the trustworthiness of long-form text generated by large language models. The ability to reliably detect nonfactual content in LLM outputs has positive societal implications, as it can help prevent the spread of misinformation and enable safer deployment of LLMs in high-stakes domains such as healthcare, law, and education. We note that our method is designed as a complementary early-warning signal rather than a replacement for comprehensive fact-checking, and is not intended for use in adversarial scenarios where models may have been deliberately trained on faulty content.

\section*{Acknowledgments}
JK thanks the Simons Foundation for support through the Collaborative Grant ``The Physics of Learning and Neural Computation''.
DB and JK acknowledge support by the NSF through NRT Award 1922658.
TGJR acknowledges support from the Foundational Research Grants program at Georgetown University's Center for Security and Emerging Technology.
This work was supported in part through the NYU IT High Performance Computing resources, services, and staff expertise.

\bibliography{references}
\bibliographystyle{include/icml/icml2026}

\clearpage

\newpage
\appendix
\onecolumn

\crefalias{section}{appsec}
\crefalias{subsection}{appsec}
\crefalias{subsubsection}{appsec}

\setcounter{equation}{0}
\renewcommand{\theequation}{\thesection.\arabic{equation}}

\small

\section*{\LARGE Appendix}
\label{sec:appendix}

\section{Full Experimental Results}
\label{sec:appendix:full_results}

\begin{table*}[h!]
\centering
\caption{\textbf{Main Experimental Results: FactScore}. Comparison of semantic isotropy with other uncertainty metrics on \textbf{FactScore-Bio} using FactScore. Each dataset column is subdivided by base model: Llama 3.1 8B, Phi 3.5 Mini, and GPT 4.1 Mini. Values are $R^2$ (explained variance) of a simple linear model of Factuality $\sim$ Isotropy score. 1,500 bootstrapped samples are used to generate 1-SD error bars. \textdagger denotes API-only models.}
\label{tab:main_results_table_fs}
\vspace{0.1in}
\begin{footnotesize}
\setlength{\tabcolsep}{10pt}
\begin{tabular}{@{}l ccc@{}}
\toprule
& \multicolumn{3}{c}{\textbf{FactScore-Bio (FactScore)}} \\
\cmidrule(lr){2-4}
\textbf{Measure / Metric}
& \textbf{Llama 3.1 8B} & \textbf{Phi 3.5 Mini} & \textbf{GPT 4.1 Mini} \\
\midrule
Perplexity \citep{ren2023outofdistributiondetectionselectivegeneration} & 0.0578   $\pm$ 0.04 & 0.0622   $\pm$ 0.04 & 0.0158   $\pm$ 0.02 \\
LN Entropy \citep{malinin2021uncertaintyestimationautoregressivestructured} & 0.0535   $\pm$ 0.04 & 0.0608   $\pm$ 0.04 & 0.00625   $\pm$ 0.009 \\
U.Eign \citep{lin2024generatingconfidenceuncertaintyquantification} & 0.344   $\pm$ 0.06 & 0.0279   $\pm$ 0.02 & 0.0185   $\pm$ 0.02 \\
U.Deg \citep{lin2024generatingconfidenceuncertaintyquantification} & 0.28   $\pm$ 0.05 & 0.0224   $\pm$ 0.02 & 0.0255   $\pm$ 0.02 \\
U.NLI \citep{manakul2023selfcheckgptzeroresourceblackboxhallucination} & 0.0916   $\pm$ 0.04 & 0.0299   $\pm$ 0.02 & 0.0263   $\pm$ 0.02 \\
EigenScore \citep{chen2024inside} & 0.107   $\pm$ 0.05 & 0.0309   $\pm$ 0.03 & N/A \\
LUQ \citep{zhang2024luq} & 0.657   $\pm$ 0.07 & 0.367   $\pm$ 0.06 & 0.192   $\pm$ 0.06 \\
Semantic Entropy \citep{Farquhar2024} & 0.136   $\pm$ 0.06 & 0.00868   $\pm$ 0.01 & 0.0152   $\pm$ 0.02 \\
\midrule
$\mathcal{I}$: Gemini\textsuperscript{\textdagger} \citep{geminiteam2024geminifamilyhighlycapable} & 0.7   $\pm$ 0.06 & 0.661   $\pm$ 0.05 & 0.48   $\pm$ 0.08 \\
$\mathcal{I}$: OpenAI Small\textsuperscript{\textdagger} \citep{gptembed} & 0.65   $\pm$ 0.06 & 0.593   $\pm$ 0.05 & 0.38   $\pm$ 0.08 \\
$\mathcal{I}$: OpenAI Large\textsuperscript{\textdagger} \citep{gptembed} & 0.649   $\pm$ 0.06 & 0.614   $\pm$ 0.05 & 0.376   $\pm$ 0.08 \\
$\mathcal{I}$: Nomic v1 \citep{nussbaum2025nomicembedtrainingreproducible} & 0.737   $\pm$ 0.05 & 0.6   $\pm$ 0.05 & 0.547   $\pm$ 0.08 \\
$\mathcal{I}$: Qwen 2 (7B) \citep{yang2024qwen2technicalreport} & 0.733   $\pm$ 0.05 & 0.589   $\pm$ 0.05 & 0.569   $\pm$ 0.08 \\
$\mathcal{I}$: SF Mistral \citep{SFRAIResearch2024} & 0.714   $\pm$ 0.05 & 0.533   $\pm$ 0.06 & \textbf{0.607   $\pm$ 0.08} \\
$\mathcal{I}$: Mistral (E5) \citep{wang2024improvingtextembeddingslarge} & 0.714   $\pm$ 0.06 & 0.508   $\pm$ 0.06 & 0.604   $\pm$ 0.08 \\
$\mathcal{I}$: Cohere v4.0 \textsuperscript{\textdagger} \citep{cohere_v4} & 0.747   $\pm$ 0.06 & \textbf{0.671   $\pm$ 0.05} & 0.5   $\pm$ 0.08 \\
$\mathcal{I}$: Cohere v3.0 \textsuperscript{\textdagger} \citep{cohere_v3} & \textbf{0.75   $\pm$ 0.06} & 0.665   $\pm$ 0.05 & 0.503   $\pm$ 0.08 \\
$\mathcal{I}$: Cohere v3.0 (Lite) \textsuperscript{\textdagger} \citep{cohere_v3} & 0.749   $\pm$ 0.05 & 0.661   $\pm$ 0.05 & 0.503   $\pm$ 0.08 \\
$\mathcal{I}$: Qwen 3 (0.6B) \citep{yang2025qwen3technicalreport} & 0.719   $\pm$ 0.06 & 0.525   $\pm$ 0.06 & 0.593   $\pm$ 0.08 \\
$\mathcal{I}$: Qwen 3 (4B) \citep{yang2025qwen3technicalreport} & 0.713   $\pm$ 0.06 & 0.588   $\pm$ 0.05 & 0.554   $\pm$ 0.08 \\
$\mathcal{I}$: Qwen 3 (8B) \citep{yang2025qwen3technicalreport}  & 0.709   $\pm$ 0.06 & 0.561   $\pm$ 0.06 & 0.528   $\pm$ 0.07 \\
\bottomrule
\end{tabular}
\end{footnotesize}
\end{table*}

\begin{table*}[h!]
\centering
\caption{\textbf{Main Experimental Results: Segment-Score}. Comparison of semantic isotropy with other uncertainty metrics across \textbf{TriviaQA Entities}, \textbf{FactScore-Bio} and \textbf{Book Summaries} datasets using the \textit{Segment-Score} algorithm. 1,500 bootstrapped samples are used to generate 1-SD error bars. \textdagger denotes API-only models. Note that EigenScore values are not generated for GPT 4.1 Mini.}
\label{tab:main_results_table_ss}
\vspace{0.1in}
\begin{scriptsize}
\setlength{\tabcolsep}{3pt}
\begin{tabular}{@{}l ccc ccc cc@{}}
\toprule
& \multicolumn{3}{c}{\textbf{TriviaQA Entities}}
& \multicolumn{3}{c}{\textbf{FactScore-Bio}}
& \multicolumn{2}{c}{\textbf{Book Summaries}} \\
\cmidrule(lr){2-4} \cmidrule(lr){5-7} \cmidrule(lr){8-9}
\textbf{Measure / Metric}
& \textbf{Llama 3.1} & \textbf{Phi 3.5} & \textbf{GPT 4.1}
& \textbf{Llama 3.1} & \textbf{Phi 3.5} & \textbf{GPT 4.1}
& \textbf{Llama 3.1} & \textbf{Phi 3.5} \\
\midrule
Perplexity \citep{ren2023outofdistributiondetectionselectivegeneration} & 0.09 $\pm$ 0.02 & 0.26 $\pm$ 0.03 & 0.00 $\pm$ 0.00 & 0.09 $\pm$ 0.04 & 0.03 $\pm$ 0.03 & 0.01 $\pm$ 0.01 & 0.01 $\pm$ 0.01 & 0.03 $\pm$ 0.02 \\
LN Entropy \citep{malinin2021uncertaintyestimationautoregressivestructured} & 0.09 $\pm$ 0.02 & 0.26 $\pm$ 0.03 & 0.03 $\pm$ 0.01 & 0.09 $\pm$ 0.04 & 0.03 $\pm$ 0.03 & 0.02 $\pm$ 0.02 & 0.01 $\pm$ 0.01 & 0.03 $\pm$ 0.02 \\
U.Eign \citep{lin2024generatingconfidenceuncertaintyquantification} & 0.06 $\pm$ 0.02 & 0.18 $\pm$ 0.03 & 0.00 $\pm$ 0.00 & 0.28 $\pm$ 0.05 & 0.05 $\pm$ 0.03 & 0.02 $\pm$ 0.02 & 0.05 $\pm$ 0.02 & 0.07 $\pm$ 0.02 \\
U.Deg \citep{lin2024generatingconfidenceuncertaintyquantification} & 0.05 $\pm$ 0.01 & 0.12 $\pm$ 0.03 & 0.00 $\pm$ 0.00 & 0.26 $\pm$ 0.07 & 0.05 $\pm$ 0.03 & 0.03 $\pm$ 0.02 & 0.06 $\pm$ 0.02 & 0.06 $\pm$ 0.03 \\
U.NLI \citep{manakul2023selfcheckgptzeroresourceblackboxhallucination} & 0.08 $\pm$ 0.02 & 0.17 $\pm$ 0.03 & 0.00 $\pm$ 0.00 & 0.14 $\pm$ 0.06 & 0.06 $\pm$ 0.03 & 0.01 $\pm$ 0.01 & 0.04 $\pm$ 0.02 & 0.06 $\pm$ 0.02 \\
EigenScore \citep{chen2024inside} & 0.02 $\pm$ 0.01 & 0.00 $\pm$ 0.00 & N/A & 0.07 $\pm$ 0.04 & 0.02 $\pm$ 0.02 & N/A & 0.02 $\pm$ 0.02 & 0.00 $\pm$ 0.01 \\
LUQ \citep{zhang2024luq} & 0.03 $\pm$ 0.01 & 0.23 $\pm$ 0.03 & 0.10 $\pm$ 0.02 & 0.38 $\pm$ 0.05 & 0.10 $\pm$ 0.05 & 0.12 $\pm$ 0.04 & 0.05 $\pm$ 0.02 & 0.33 $\pm$ 0.04 \\
Semantic Entropy \citep{Farquhar2024} & 0.03 $\pm$ 0.01 & 0.08 $\pm$ 0.02 & 0.00 $\pm$ 0.00 & 0.16 $\pm$ 0.07 & 0.03 $\pm$ 0.02 & 0.01 $\pm$ 0.01 & 0.02 $\pm$ 0.01 & 0.03 $\pm$ 0.02 \\
\midrule
$\mathcal{I}$: Gemini\textsuperscript{\textdagger} \citep{geminiteam2024geminifamilyhighlycapable} & \textbf{0.24 $\pm$ 0.03} & \textbf{0.43 $\pm$ 0.02} & 0.03 $\pm$ 0.01 & 0.44 $\pm$ 0.04 & 0.24 $\pm$ 0.07 & 0.34 $\pm$ 0.04 & 0.26 $\pm$ 0.03 & \textbf{0.43 $\pm$ 0.04} \\
$\mathcal{I}$: OpenAI Small\textsuperscript{\textdagger} \citep{gptembed} & 0.16 $\pm$ 0.02 & 0.28 $\pm$ 0.02 & 0.00 $\pm$ 0.00 & 0.44 $\pm$ 0.04 & 0.27 $\pm$ 0.07 & 0.31 $\pm$ 0.04 & 0.26 $\pm$ 0.03 & 0.35 $\pm$ 0.03 \\
$\mathcal{I}$: OpenAI Large\textsuperscript{\textdagger} \citep{gptembed} & 0.16 $\pm$ 0.02 & 0.27 $\pm$ 0.02 & 0.00 $\pm$ 0.00 & 0.44 $\pm$ 0.03 & 0.27 $\pm$ 0.06 & 0.32 $\pm$ 0.04 & 0.20 $\pm$ 0.03 & 0.37 $\pm$ 0.03 \\
$\mathcal{I}$: Nomic v1 \citep{nussbaum2025nomicembedtrainingreproducible} & 0.27 $\pm$ 0.02 & 0.36 $\pm$ 0.02 & \textbf{0.14 $\pm$ 0.03} & 0.45 $\pm$ 0.04 & 0.27 $\pm$ 0.06 & 0.42 $\pm$ 0.04 & 0.24 $\pm$ 0.03 & 0.34 $\pm$ 0.03 \\
$\mathcal{I}$: Qwen 2 (7B) \citep{yang2024qwen2technicalreport} & 0.08 $\pm$ 0.02 & 0.27 $\pm$ 0.03 & 0.02 $\pm$ 0.01 & 0.49 $\pm$ 0.04 & \textbf{0.36 $\pm$ 0.06} & 0.42 $\pm$ 0.04 & 0.12 $\pm$ 0.03 & 0.30 $\pm$ 0.03 \\
$\mathcal{I}$: SF Mistral \citep{SFRAIResearch2024} & 0.08 $\pm$ 0.02 & 0.30 $\pm$ 0.03 & 0.01 $\pm$ 0.01 & 0.45 $\pm$ 0.04 & 0.29 $\pm$ 0.06 & 0.46 $\pm$ 0.04 & 0.11 $\pm$ 0.02 & 0.24 $\pm$ 0.03 \\
$\mathcal{I}$: Mistral (E5) \citep{wang2024improvingtextembeddingslarge} & 0.11 $\pm$ 0.02 & 0.32 $\pm$ 0.02 & 0.00 $\pm$ 0.00 & 0.45 $\pm$ 0.04 & 0.28 $\pm$ 0.06 & \textbf{0.47 $\pm$ 0.04} & 0.09 $\pm$ 0.02 & 0.23 $\pm$ 0.03 \\
$\mathcal{I}$: Cohere v4.0 \textsuperscript{\textdagger} \citep{cohere_v4} & 0.11 $\pm$ 0.02 & 0.28 $\pm$ 0.03 & 0.00 $\pm$ 0.00 & \textbf{0.50 $\pm$ 0.03} & 0.30 $\pm$ 0.07 & 0.41 $\pm$ 0.04 & 0.24 $\pm$ 0.03 & 0.41 $\pm$ 0.03 \\
$\mathcal{I}$: Cohere v3.0 \textsuperscript{\textdagger} \citep{cohere_v3} & 0.11 $\pm$ 0.02 & 0.28 $\pm$ 0.03 & 0.00 $\pm$ 0.00 & 0.50 $\pm$ 0.04 & 0.27 $\pm$ 0.07 & 0.41 $\pm$ 0.04 & 0.20 $\pm$ 0.03 & 0.34 $\pm$ 0.03 \\
$\mathcal{I}$: Cohere v3.0 (Lite) \textsuperscript{\textdagger} \citep{cohere_v3} & 0.11 $\pm$ 0.02 & 0.28 $\pm$ 0.03 & 0.00 $\pm$ 0.00 & 0.50 $\pm$ 0.03 & 0.24 $\pm$ 0.06 & 0.41 $\pm$ 0.04 & 0.25 $\pm$ 0.03 & 0.36 $\pm$ 0.03 \\
$\mathcal{I}$: Qwen 3 (0.6B) \citep{yang2025qwen3technicalreport} & 0.05 $\pm$ 0.02 & 0.17 $\pm$ 0.02 & 0.01 $\pm$ 0.01 & 0.47 $\pm$ 0.04 & 0.31 $\pm$ 0.06 & 0.42 $\pm$ 0.04 & 0.18 $\pm$ 0.03 & 0.40 $\pm$ 0.03 \\
$\mathcal{I}$: Qwen 3 (4B) \citep{yang2025qwen3technicalreport} & 0.01 $\pm$ 0.01 & 0.17 $\pm$ 0.03 & 0.14 $\pm$ 0.03 & 0.44 $\pm$ 0.04 & 0.31 $\pm$ 0.06 & 0.37 $\pm$ 0.05 & 0.31 $\pm$ 0.03 & 0.40 $\pm$ 0.03 \\
$\mathcal{I}$: Qwen 3 (8B) \citep{yang2025qwen3technicalreport}  & 0.00 $\pm$ 0.00 & 0.12 $\pm$ 0.02 & 0.13 $\pm$ 0.03 & 0.43 $\pm$ 0.04 & 0.28 $\pm$ 0.06 & 0.33 $\pm$ 0.05 & \textbf{0.33 $\pm$ 0.03} & 0.36 $\pm$ 0.04 \\
\bottomrule
\end{tabular}
\end{scriptsize}
\end{table*}

\begin{table}[h!]
\centering
\newcommand{\spm}{\ensuremath{\mathbin{\scriptstyle\pm}}}
\captionsetup{width=\linewidth}
\caption{\small \textbf{Comparison of different isotropy measures}: Analysis of various isotropy measures. Comparing explained variance ($\smash{R^2}$) of Factuality on $\approx$500 word length responses generated using Phi 3.5 Mini Instruct on the \textbf{TriviaQA} \textbf{(SS)} dataset. 1,500 bootstrapped samples are used to generate 1-SD error bars.}
\vspace{-5pt}
\label{tab:isotropy-measures}
\vspace{0.1in}
\begin{small}
\setlength{\tabcolsep}{28.5pt}
\begin{tabular}{lcccc}
\toprule
\textbf{Model} & \textbf{Frob.} & \textbf{Inv.\ Tr.} & \textbf{LogDet} & \textbf{\textrm{vNE}} \\
\midrule
Gemini   & 0.41 \spm 0.02 & 0.38 \spm 0.02 & \textbf{0.43 \spm 0.02} & 0.43 \spm 0.02 \\
Nomic v1 & 0.33 \spm 0.02 & 0.33 \spm 0.02 & \textbf{0.39 \spm 0.02} & 0.35 \spm 0.02 \\
OpenAI (L)  & 0.26 \spm 0.02 & 0.21 \spm 0.02 & 0.26 \spm 0.02 & \textbf{0.27 \spm 0.02} \\
OpenAI (S)  & 0.27 \spm 0.02 & 0.17 \spm 0.02 & 0.24 \spm 0.02 & \textbf{0.28 \spm 0.02} \\
Cohere v4 & 0.27 \spm 0.02 & 0.22 \spm 0.02 & 0.26 \spm 0.02 & \textbf{0.28 \spm 0.02} \\
Cohere v3 & 0.27 \spm 0.02 & 0.22 \spm 0.02 & 0.26 \spm 0.02 & \textbf{0.28 \spm 0.02} \\
Qwen 2   & \textbf{0.27 \spm 0.02} & 0.17 \spm 0.02 & 0.23 \spm 0.02 & 0.27 \spm 0.02 \\
Mistral  & 0.29 \spm 0.02 & 0.19 \spm 0.02 & 0.27 \spm 0.02 & \textbf{0.30 \spm 0.02} \\
\bottomrule
\end{tabular}
\end{small}
\end{table}

\section{Experimental Details}

\begin{wrapfigure}{r}{0.5\textwidth}
    \centering
    \vspace{-20pt}
    \includegraphics[width=\linewidth,trim={0 25 160 20},clip]{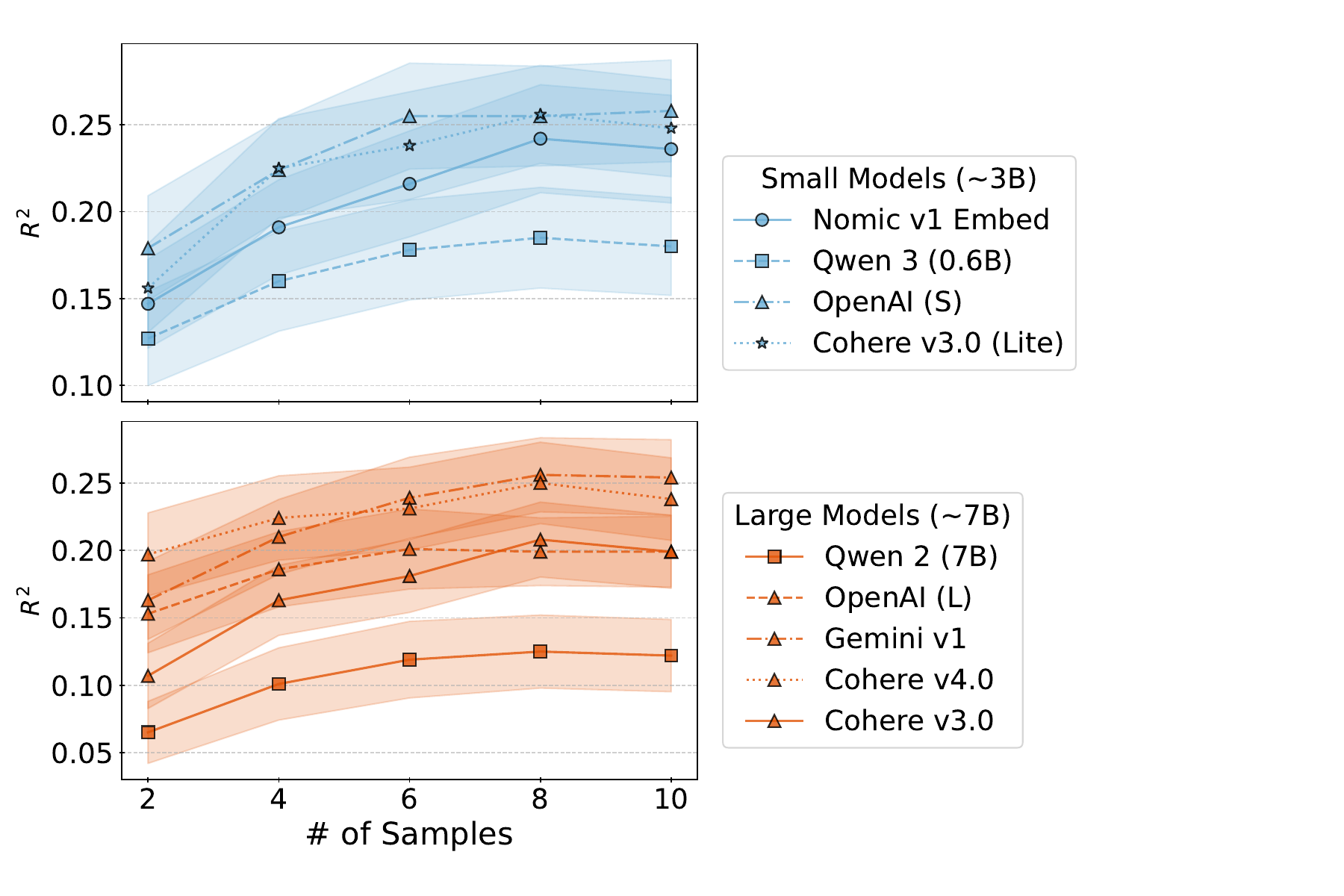}
    
    \captionsetup{width=\linewidth}
    
    \caption{\textbf{Performance by \# of Samples}: Line Charts comparing the performance of semantic isotropy by number of sampled responses on \textbf{FS-BIO (SS)} dataset of $\approx\hspace*{-3pt}500$ words, contrasted by model size: Small (up to $3B$ parameters) and Large ($\approx\hspace*{-3pt}7B$ or more parameters). 1,500 bootstrapped samples are used to generate 1-SD error regions.}
    \label{fig:number_of_samples_analysis}
    \vspace{-20pt}
\end{wrapfigure}

\subsection{Effect of Number of Sampled Responses.}

\Cref{fig:number_of_samples_analysis} illustrates how performance scales with the number of sampled responses. Encouragingly, semantic isotropy scoring does not require many samples ($\approx\hspace*{-3pt}6$-$8$) to achieve comparable performance to its implied asymptotic level (at $N=10$ samples), across both small and large models.
While the method is generally scalable, this result confirms that it can deliver consistent truthfulness indicators with minimal overhead.

\subsection{Effect of Generator Capability Level.}
In addition to the results discussed in \Cref{sec:main_results}, where we used long-form responses generated by Phi 3.5 Mini Instruct and Llama 3.1 8B Instruct, we also considered long-form responses generated by a more capable, closed-weight model, GPT 4.1 Mini.
Applying the {\it Segment-Score} method to long-form responses for FS-BIO and TriviaQA, we find that the absolute performance of semantic isotropy scoring as well as its relative performance compared to the baselines significantly increases on FS-BIO but decreases on TriviaQA.
On TriviaQA, semantic isotropy scoring achieves a higher $R^2$ than all baselines using Nomic v1 embeddings, but the best absolute $R^2$ value is barely above 10\%, implying that almost none of the variation in the factuality scores can be explained by any of the methods used.
Since, for FS-BIO, semantic isotropy scoring performs better on long-form responses generated by GPT 4.1 Mini than on long-form responses generated by Phi 3.5 Mini Instruct and Llama 3.1 8B Instruct, the poor performance (across methods) on TriviaQA is unlikely to be due to increased model capability per se and more likely due to dataset idiosyncrasies, such as inherent entity ambiguities, or model pre-training artifacts.
We present a detailed analysis and discussion in \Cref{sec:openai_gpt_4.1_mini_ablation}.

\subsection{Computational Considerations.}
\label{sec:appendix:compute}
End-to-end timing for the proposed method versus the strongest NLI-based baseline is reported in \Cref{sec:compute}. We summarize the asymptotic analysis here for completeness. Existing methods such as LUQ and Graph Uncertainty scale as $\mathbf{O}(MN^2)$, where $M$ is the average number of facts or segments per response and $N$ is the number of sampled responses, since each pair requires a separate NLI forward pass. Semantic entropy and derived methods are $\mathbf{O}(N^2)$ in entailment computations, which also cannot be naively vectorized because the NLI model consumes (premise, hypothesis) string pairs rather than tensors. In contrast, while semantic isotropy is naively $\mathbf{O}(N^2)$, the cosine kernel can be computed in a fully parallelized fashion, yielding substantial performance improvements.

\subsection{Effect of Choice of Embedding Aggregation.} 
\begin{figure*}
  \centering
  \includegraphics[width=1\textwidth]{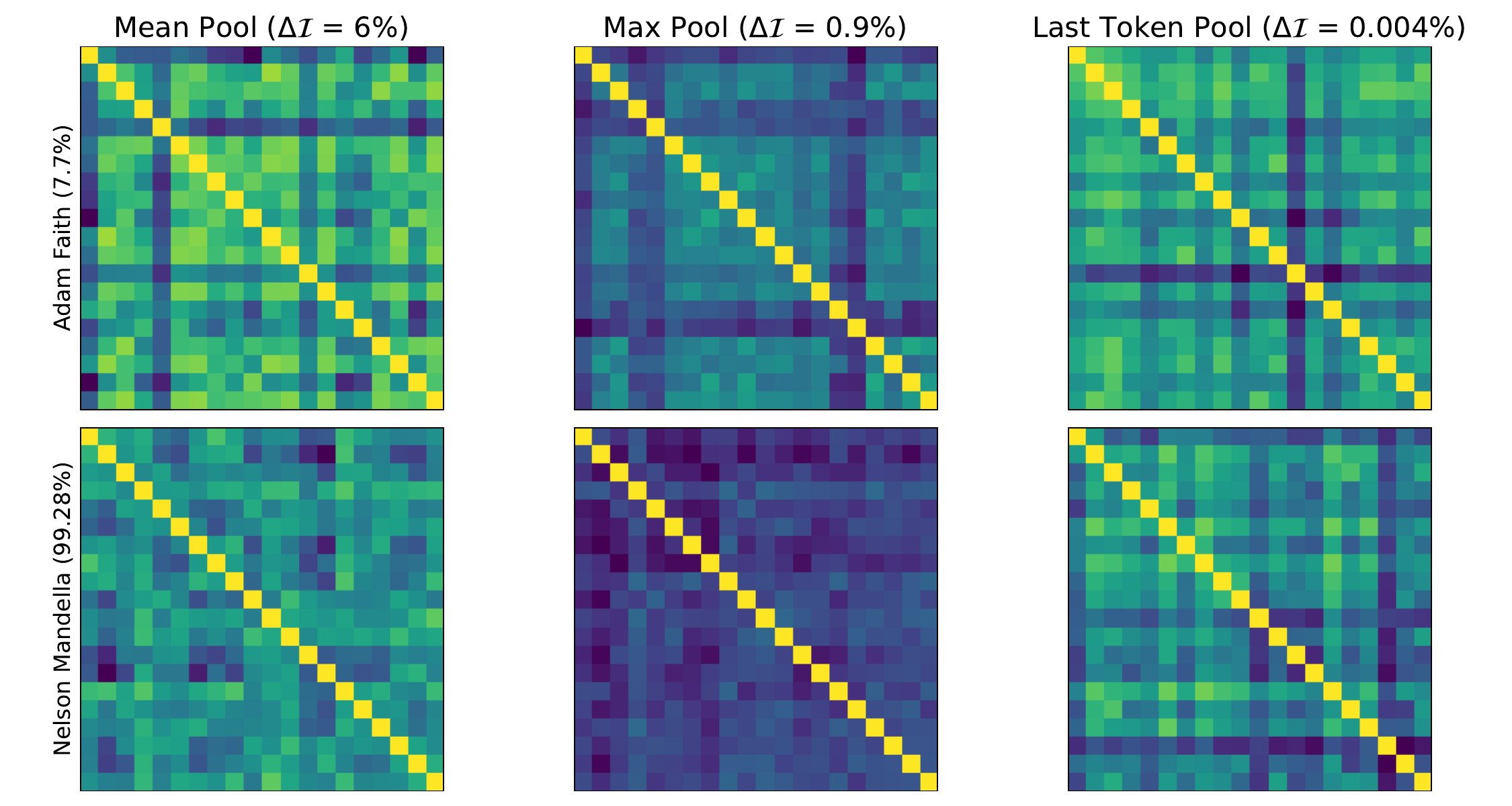}
  \caption{\textbf{Nomic v1 Embedding Choice}. $K_E^{\textrm{\em cos}}$ for two sample topics: \textbf{Adam Faith} (low factuality, top row) and \textbf{Nelson Mandela} (high factuality, bottom row). The number in the parenthesis next to the topic name is the average factuality across sampled responses. $\Delta \mathcal{I}$ for each pooling method denotes the percentage difference in semantic isotropy scores between the two topics (as a percentage of $\mathcal I$ for \textbf{Nelson Mandela}). We infer that Mean Pooling for Nomic v1 yields the most informative matrices for an isotropy analysis.}
  \label{fig:nomic_emb}
\end{figure*}

Our experimental results require an embedding vector $z_i \in \mathbb{R}^D$. Closed-weight models return a single vector per response and allow the user to configure embedding size ($D$) as an input. We use the model default for Gemini (768), OpenAI (1536) and Cohere (1024). However, Open-weight embedding models like Qwen 2 \citep{yang2024qwen2technicalreport} and Mistral \citep{SFRAIResearch2024} specify using the last token embedding in the final hidden state. This can be found in their reference implementations on \href{https://huggingface.co/Alibaba-NLP/gte-Qwen2-7B-instruct}{HuggingFace}.
In contrast, for Nomic v1 \citep{nussbaum2025nomicembedtrainingreproducible}, authors do not recommend a pooling method and encourage users to find methods that work well for their use case. We find that the last token embedding method yields a weak semantic isotropy score. Instead, we use the mean pooling over the token dimension. Given model activations of the last hidden layer as a matrix $E \in \mathbb{R}^{L \times D}$, where $L$ is the number of tokens in the response and $D$ the embedding dimension, we produce $Z_{\text{\texttt{nomic}}} = \texttt{mean(}E, \texttt{dim=1})$ such that $Z_{\text{\texttt{nomic}}} \in \mathbb{R}^{D}$. In \Cref{fig:nomic_emb}, we show the difference between a highly trustworthy and a low trustworthy topic by mean, max and last token pooling for Nomic v1.

\subsection{Infrastructure Requirements}
\label{sec:appendix:infra}
Our work was performed using Google Cloud Platform compute to run inference on the white-box models using a single node with up to 4 NVIDIA Tesla V100 GPUs. GPT 4.1 Mini was used as the oracle LLM in Algorithm \ref{alg:segment_verify}. OpenAI \texttt{text-embedding-3-small},  \texttt{text-embedding-3-large}, Gemini \texttt{gemini-embedding-001}, Cohere Embedding v4.0, v3.0 \citep{cohere_v3, cohere_v4} were also used to compute response embeddings for semantic isotropy  computation. Additionally, OpenAI GPT 4.1, Anthropic Claude 4 Sonnet, and  DeepSeek \citep{deepseekai2024deepseekv2strongeconomicalefficient} were also used in verification runs to study robustness of the scoring pipeline.
Overall, we utilize approximately 1,440 GPU hours (in addition to CPU compute time) to generate samples and score the responses.

\subsection{Deploying Semantic Isotropy Score in a Task-Specific Context}
\label{sec:si_task}

The discriminative analysis is presented in the main body as \Cref{sec:auroc} and \Cref{fig:auroc}. As measures of utility or correctness over long-form generations could be real valued and non-binary, we discretize the experimental setup by classifying responses with average factuality $> 50\%$ as $1$ and $0$ otherwise, then use $1-\mathcal{I}$ as the confidence measure. The resulting plots are consistent with our regression results, with performance varying across datasets while working well in most settings; in the more specific datasets such as \textbf{FS-BIO} and \textbf{BS}, the score performs particularly well at classifying responses.

\subsection{Segment-Score Implementation Specifics}
\label{sec:segscore_method}
Unlike \textit{FactScore}, the \textit{Segment-Score} algorithm is achieved through a single prompt per response. The basic prompt template for our method can be found in \Cref{fig:prompt_instructions_template}. The prompt is structured into three components. First, an instruction block that covers the overall objective of the scoring exercise, expected data, desired output structure, and specific handling for edge cases. Next, we provide  two manually curated examples to guide the model's output. One such example can be found in \Cref{fig:segscore_prompt_example}.
Last, we provide the entity name, contents of the reference document and response to be scored. The model is prompted to respond using an XML tag structure which simplifies the data extraction process. We also force the model to respond with only \texttt{`0'} or \texttt{`1'} for the output classes. For 99.8\% of segments, the top 2 tokens are the `0' or `1' class labels, and we can use the log-probabilities from the API to generate a normalized probability score for each labeled segment.

\section{Ablation Studies}

\subsection{Response Generation Using OpenAI GPT 4.1 Mini}
\label{sec:openai_gpt_4.1_mini_ablation}

\begin{figure*}[t!]
  \centering
  \includegraphics[width=1\textwidth,trim={0 25 0 15},clip]{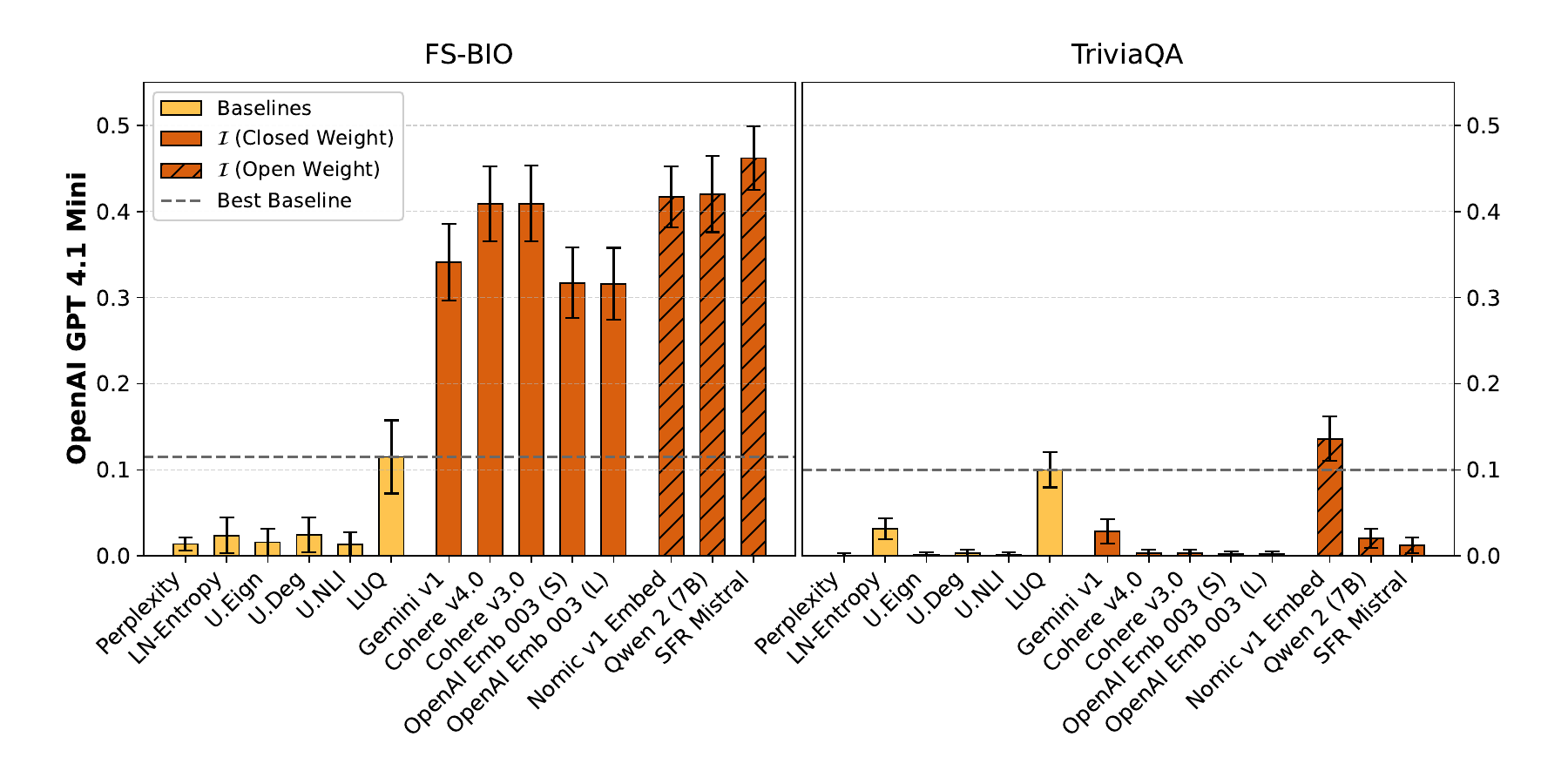}
  \vspace*{-15pt}
    \caption{\textbf{Experimental Results using OpenAI GPT 4.1 Mini}. Bar charts comparing the performance (as measured by the $R^2$ of a linear model of factuality with semantic isotropy as the explanatory variable); implemented using various embedding models and benchmark uncertainty metrics on both the \textbf{FS-BIO} and \textbf{TriviaQA} datasets using the \textit{Segment-Score} \textbf{(SS)} scoring algorithm. Left hand side: Benchmark UQ metrics. Right Hand Side: semantic isotropy  (\textbf{$\mathcal{I}$}) implemented using various embedding models. Same experimental setting as in \Cref{fig:main_results}. We observe that performance on generations using OpenAI GPT 4.1 Mini is significantly lower than other generation models; however, this extends to all benchmark metrics as well. Semantic isotropy continues to outperform all other baselines. Note that EigenScore values are not generated for GPT 4.1 Mini as it is a black-box model and the internal layer activations are not available.}
    \label{fig:openai_analysis}
    \vspace*{-10pt}
\end{figure*}

To understand the impact of the generator model on the performance of the semantic isotropy score, we also generate responses using OpenAI GPT 4.1 Mini \citep{gpt41}, scored using \textit{Segment-Score}. \Cref{fig:openai_analysis} shows the results of this experiment in the same format as our main results. For the \textbf{FS-BIO} dataset, we see semantic isotropy remains predictive of overall response factuality. However, we see a notable drop in performance in the case of \textbf{TriviaQA}. Notably, this extends to all benchmark metrics as well. We also observe much greater variance in the performance across embedding models, suggesting there are some settings where the choice of embedding model is crucial.

Consequently, it was important to understand the impact of the generation model on the performance of the semantic isotropy score. \citet{zhang2024luq} note that the LUQ method does not perform well using GPT 4, a fact we were able to replicate.

\begin{quote}
\textit{"We also observe that LUQ is better suited for models with relatively lower factuality
and a lack of self-expressiveness regarding uncertainty. For models with high factuality capabilities,
such as GPT-4, LUQ only demonstrates a moderate correlation with factuality scores"} - \cite{zhang2024luq}
\end{quote}

\begin{figure}[t!]
    \centering
    \begin{subfigure}[t]{0.48\textwidth}
          \centering
          \includegraphics[width=1\textwidth,trim={0 15 0 15},clip]{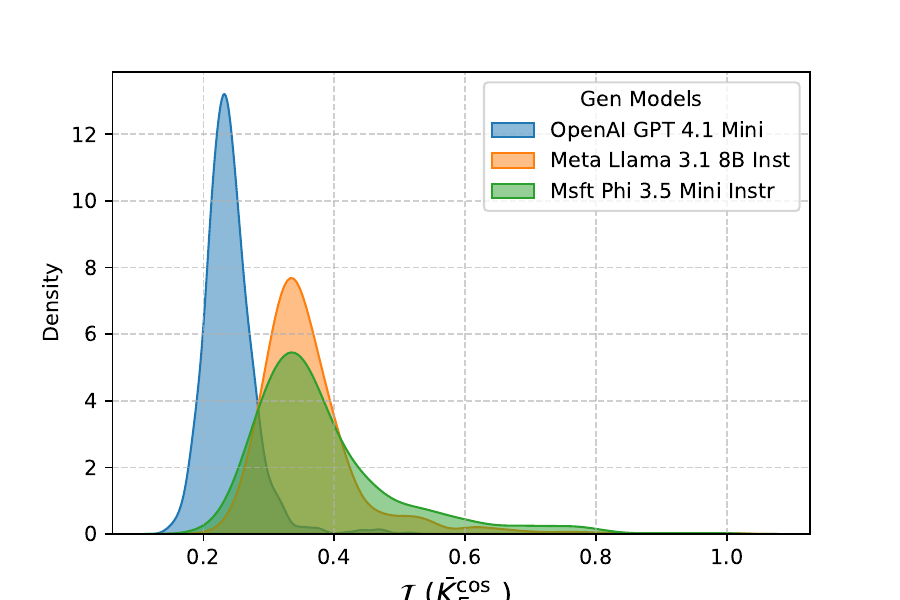}
          \vspace*{-15pt}
            \caption{\textbf{Kernel Density Estimation plots of semantic isotropy scores for TriviaQA}. Kernel Density Estimation plots of semantic isotropy scores for TriviaQA, comparing the distributions of scores for GPT 4.1 Mini, Phi 3.5 Mini and Llama 3.1 8B. The embedding used in Gemini v1 \citep{geminiteam2024geminifamilyhighlycapable}. We observe that GPT 4.1 Mini's scores are smaller in magnitude and more tightly clustered.}
            \label{fig:isotropy_kde}
    \end{subfigure}
    \hfill
    \begin{subfigure}[t]{0.48\textwidth}
        \centering
        \includegraphics[width=1\linewidth]{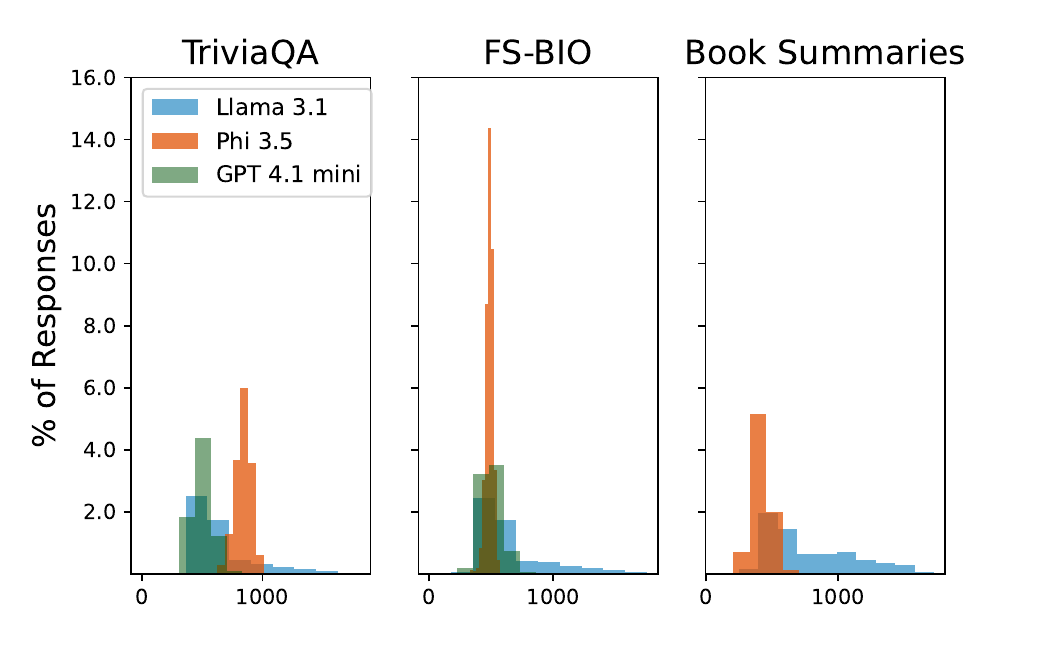}

        \captionsetup{width=.95\linewidth}

        \caption{\textbf{Word Count by Generation Model and Dataset}: Density histograms comparing the response length (in tokens) for each generative model and dataset for a prompt with a guidance length of 500 words. We observe that GPT 4.1 Mini is much better at instruction following than Llama 3.1 and Phi 3.5 Mini, sticking fairly closely to the intended word limit.}
        \label{fig:token_count_distribution}
    \end{subfigure}
    \caption{\small \textbf{Response Characteristics:} Comparing the characteristics of generated responses by model type and dataset. Distribution of semantic isotropy scores for TriviaQA using Gemini [Left] and Histogram of response lengths in tokens by dataset and generative model [Right].}
  \vspace*{-10pt}
\end{figure}

One would expect that GPT 4.1 Mini would perform better in terms of instruction following and maintaining consistent structure. This is anecdotally observed in the token count distribution (see \Cref{fig:token_count_distribution}), where GPT 4.1 Mini adheres much better to the 500 word guidance given in the prompt across different topic areas, as compared to the other generative models being considered.

When studying the distribution of semantic isotropy scores, in \Cref{fig:isotropy_kde}, we observe that OpenAI GPT 4.1 Mini's scores are smaller in magnitude and more tightly clustered. This highlighted that there may be some inherent structural properties of the OpenAI GPT 4.1 Mini generations that manifested in the embeddings in this specific way. To tackle this, we performed a comparative study between generations from GPT 4.1 Mini \citep{gpt41}, Phi 3.5 Mini \citep{abdin2024phi3technicalreporthighly} and Llama 3.1 8B \citep{grattafiori2024llama3herdmodels}. Using an LLM-as-a-judge setup, we asked OpenAI GPT 4.1 to compare and contrast 10 generations from GPT 4.1 Mini vs 10 from either Phi 3.5 Mini or Llama 3.1 8B for 50 different topics where all of the models had relatively low factuality (i.e. $\leq 33\%$ accuracy on average). The LLM generated comparisons were summarized into a brief analysis (See \Cref{fig:openai_comparative_analysis_summary}). This highlighted that properties such as structure, tone and other qualitative aspects of a generation materially impacted the resulting embeddings. We expect that improvements in embedding models or models specifically tuned to disregard qualitative aspects of a generation could improve the performance of the semantic isotropy score, but leave this exploration for future work.

\begin{figure*}[t]
    \centering
    \begin{lstlisting}[linewidth=\textwidth]
      Across the comparisons, Model A consistently demonstrates formal, polished, and structured writing with high factual accuracy,  depth, and analytical rigor. Its responses are typically essay-like, varied in perspective, and nuanced, often employing advanced vocabulary, rhetorical devices, and critical engagement with broader themes such as legacy, cultural context, and impact.

      In contrast, Model B tends to be more informal, conversational, and formulaic, frequently exhibiting repetition, superficial coverage, and factual errors. Model B's diversity often arises from inconsistency or error rather than true creativity, and its outputs are generally less coherent, less organized, and less engaging than Model A's.
    \end{lstlisting}
    \caption{Summary Analysis of Low Factuality Generations by GPT 4.1 Mini (Model A) vs Llama 3.1 8B (Model B). }
    \label{fig:openai_comparative_analysis_summary}
\end{figure*}

\subsection{Comparing the Efficiency of the Two Scoring Methods}
\label{sec:ss_vs_fs}

{\em Segment-Score's} overall process closely follows the FactScore algorithm. However, we make several key optimizations that improve runtime. First, unlike FactScore, we break the input text into atomic segments rather than extracting atomic facts. As a result, each sentence may be broken into 2-3 segments that need to be scored whereas FactScore may generate several atomic facts pertaining to each sentence segment. For \textbf{FS-BIO}, this translates to $28~\pm~6$ distinct segments for each response in \textit{Segment-Score}, while \textit{FactScore} produces $100 ~ \pm ~ 20$ atomic facts (2-$\sigma$ interval).

\begin{figure*}[h!]
  \centering
  \includegraphics[width=0.9\textwidth]{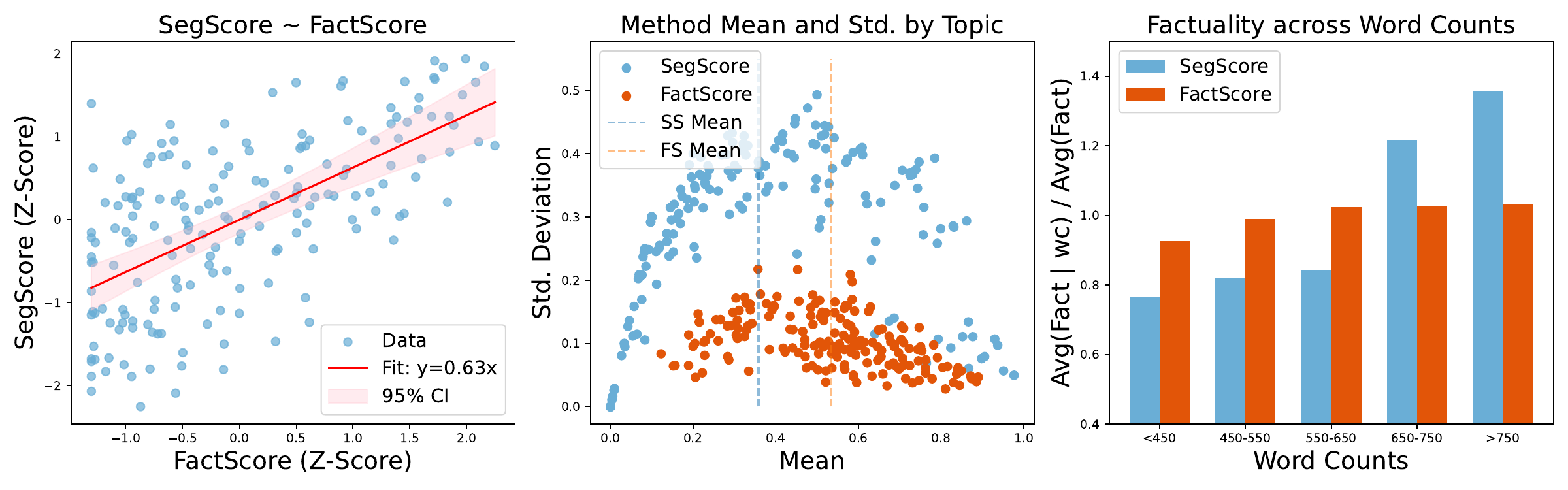}
  \caption{\textbf{FS vs SS [FS-BIO]}. Comparison of \textit{FactScore} and \textit{Segment-Score} on the \textbf{FS-BIO} dataset. \textit{Segment-Score} yields a dataset with lower overall factuality and higher variance; however, the relationship between the two scores is robust and consistent.}
  \label{fig:fsbio_fs_v_ss}
  \vspace*{-10pt}
\end{figure*}

Second, FactScore does not distinguish between verifiable and ambiguous facts which increases noise in the final results. We explicitly prompt the model to classify such examples as False. This generally biases our result lower, which can be seen in the middle plot of \Cref{fig:fsbio_fs_v_ss}.

Finally, our entire process is achieved using one single query to the oracle, while FactScore may use an order of magnitude more depending on the number of atomic facts extracted and the length of the ground truth document. Consequently, our method scales more favorably as the length of the response to be scored increases. Despite these differences, we see comparable results for \textbf{FS-BIO} in \Cref{tab:main_results_table_fs} and \Cref{tab:main_results_table_ss} for each scoring method when considering the various isotropy configurations and response generating models.

\section{Extended Discussion of Limitations}

A condensed version of this discussion appears in the main body (\Cref{sec:limitations}). The first drawback of our method is the need for repeated sampling, which can be expensive to generate.
Approaches such as \citet{entropyprobe} have shown that it is possible to train cheap probes that can a priori estimate uncertainty scores without needing repeated sampling. With the advent of reasoning models with long Chain-of-Thought reasoning contexts \citep{zhao2024marcoo1openreasoningmodels} and Monte-Carlo Tree Search \citep{wang2024selfimprovementllmsmctsleveraging}, as well as fast inference engines with significantly higher throughput \citep{kwon2023efficientmemorymanagementlarge}, the relative overhead of repeated sampling has been greatly diminished, making methods like ours compelling.

Additionally, while studies such as \citet{manakul2023selfcheckgptzeroresourceblackboxhallucination} have observed that hallucinated responses tend to diverge, and generally are inconsistent with each other, \citet{ricco2025hallucinationdetectionprobabilisticframework} find that this effect is not universal and introduce the taxonomy of \textit{extrinsic} (characterized by arbitrary and orthogonal content) vs \textit{intrinsic} (contradictory or anti-parallel content) hallucination. The semantic isotropy  framework would likely fail in cases of \textit{intrinsic} hallucinations. However, in practice these intrinsic hallucinations would be exceedingly rare, especially in the long-form response format.

Finally, it is important to call out that our method is not designed to work in adversarial scenarios. In the event that one were to use \textit{semantic isotropy} on a model that had been intentionally trained on faulty content, semantic consistency across samples would only indicate the model is faithfully inferring from its training corpora, and would not indicate factuality or trustworthiness. While \textit{semantic isotropy} remains far more robust to naive manipulation techniques such as ballot stuffing, whereby facts are repeatedly stated to boost accuracy metrics, this failure case is beyond the scope of standard embedding models. In this case, we deem \textit{semantic isotropy} as a complementary early-warning signal, rather than a full-fledged replacement for claim-by-claim verification.

\section{Reproducibility}

We have made a significant effort to ensure the reproducibility of our results.
Code is provided at the GitHub repository linked in the introduction, which includes training, evaluation, and analysis scripts.
The experimental setup---including hyperparameters, model configurations, and sampling parameters---is described in \Cref{sec:dataset_generation}.
All datasets used in our experiments are publicly available, and we additionally provide scripts for data preparation.
Our experiments rely on open-weight language models and certain API-only models; instructions for reproducing results with both are included in the \texttt{README.md} file in the released code.
Infrastructure and compute requirements are documented in \Cref{sec:appendix:infra}.
The {\it Segment-Score}-annotated dataset is released alongside the code repository, enabling direct benchmarking of our results.

\lstset{
    basicstyle=\ttfamily\small,
    breaklines=true,
    breakatwhitespace=false,
    frame=single,
    showstringspaces=false,
    columns=flexible,
    captionpos=b
}

\begin{figure*}[t]
    \centering
    \begin{lstlisting}[linewidth=\textwidth]
// INSTRUCTION PROMPT FOR SEGMENT SCORE
You are an NLP segmentation and evaluation engine.
Examine the scenario below. You are given:
1. The name of an entity/person/place/thing etc. in <entity> tags.
2. A reference document regarding the entity in <reference_doc> tags.
3. A response about the entity to evaluate in <response> tags.

### Your Tasks:
1. **Segmentation Task:**
   Segment the `<response>` into individual statements. Each statement can be a sentence, phrase or word and should convey a single, complete, and independent piece of information about the `<entity>`. Do not modify, rephrase, or paraphrase the original text. Ensure no semantic overlaps exist between statements. Individual proper nouns should be part of their own statement however when determining the appropriate classification, preceding context can be used when appropriate.
   Verify that the concatenated content of all statements exactly matches the original response.

   Format the segmented response as follows:
   ```
   <statements>
   <statement>Statement 1</statement> <class>Class 1</class>
   <statement>Statement 2</statement> <class>Class 2</class>
   ...
   </statements>
   ```

2. **Factual Classification Task:**
   For each segmented `<statement>`, classify it as 1 (True) or 0 (False) based solely on the information in the `<reference_doc>`. Follow these guidelines:
   - If a statement is factually accurate and supported by the `<reference_doc>`, classify it as '1'.
   - If a statement is inaccurate, unverifiable, or not supported by the `<reference_doc>`, classify it as '0'.
   - If a statement is partially true, but contains incorrect or unsupported information, classify it as '0'.
   - Do not rely on any external knowledge or context beyond the `<reference_doc>`.
   - Include only the classification for each statement. Do not provide any explanations or additional information.
   - Specify the class in <class> tags.
   - The ONLY valid class values are `1` and `0`. No other values or words should appear within the `<class>` tags.

3. **Error Handling:**
   - If the `<response>` contains unparseable text, incomplete sentences, or conflicting information that cannot be resolved using the `<reference_doc>`, include the flagged statement as is and classify it as 'False'.

Examples:
####### EXAMPLE 1 ######
{{example_one}}
########################
####### EXAMPLE 2 ######
{{example_two}}
########################

Entity:
<entity>
{entity}
</entity>

Reference Document:
<reference_doc>
{reference_doc}
</reference_doc>

Response to Evaluate:
<response>
{response}
    \end{lstlisting}
    \caption{Segment-Score Prompt Template with instructions}
    \label{fig:prompt_instructions_template}
\end{figure*}

\begin{figure*}[t]
    \centering
    \begin{lstlisting}[linewidth=\textwidth]
<entity>
London, UK
</entity>

Reference Document:
<reference_doc>
London, England's capital, boasts a rich history spanning millennia. Founded by the Romans as Londinium around 47 AD, it became a major port and trading center. After the Roman withdrawal, Anglo-Saxons established Lundenwic, which later fell to Viking raids. The Norman Conquest in 1066 led to the construction of the Tower of London, a symbol of royal power. London thrived during the medieval period, becoming a major center for trade, finance, and culture. It weathered plagues, fires, and civil wars, emerging as a global metropolis and the heart of the British Empire. Today, London remains a vibrant hub, blending its historical legacy with modern dynamism, home to over 9 million people.
</reference_doc>

Response to Evaluate:
<response>
London, the capital city of England and the United Kingdom, is a vibrant metropolis steeped in history and brimming with modern energy. With a population of over 9 million people, it stands as one of the world's most influential global cities, known for its diverse culture, iconic landmarks, and rich heritage.

The city's history stretches back over three millennia, founded by the Romans as Londinium in 43 AD. Throughout the centuries, London has played a pivotal role in world affairs, serving as the heart of the British Empire and surviving tumultuous events such as the Great Fire of 1666 and the Blitz during World War I.

Today, London is a melting pot of cultures, with over 300 languages spoken within its boundaries. This diversity is reflected in its neighborhoods, each with its own unique character and charm. From the trendy streets of Shoreditch to the upscale boutiques of Mayfair, there's something for everyone in this cosmopolitan city.
</response>

Segmented and classified response:
<statements>
<statement>London, the capital city of England and the United Kingdom</statement> <class>1</class>
<statement>is a vibrant metropolis steeped in history and brimming with modern energy</statement> <class>1</class>
<statement>With a population of over 9 million people</statement> <class>1</class>
<statement>it stands as one of the world's most influential global cities, known for its diverse culture, iconic landmarks, and rich heritage.</statement> <class>1</class>
<statement>The city's history stretches back over three millennia, founded by the Romans as Londinium in 43 AD</statement> <class>0</class>
<statement>Throughout the centuries, London has played a pivotal role in world affairs, serving as the heart of the British Empire</statement> <class>1</class>
<statement>and surviving tumultuous events such as the Great Fire of 1666</statement> <class>1</class>
<statement>and the Blitz during World War I</statement> <class>0</class>
<statement>Today, London is a melting pot of cultures, with over 300 languages spoken within its boundaries</statement> <class>1</class>
<statement>This diversity is reflected in its neighborhoods, each with its own unique character and charm.</statement> <class>1</class>
<statement>From the trendy streets of Shoreditch to the upscale boutiques of Mayfair, there's something for everyone in this cosmopolitan city</statement> <class>1</class>
</statements>
    \end{lstlisting}
    \caption{Segment-Score Example used as part of model in-context training for oracle LLM.}
    \label{fig:segscore_prompt_example}
\end{figure*}

\end{document}